\documentclass{article}

\usepackage[final]{neurips_2024}

\usepackage[utf8]{inputenc} 
\usepackage[T1]{fontenc}    
\usepackage[colorlinks]{hyperref}       
\usepackage{url}            
\usepackage{booktabs}       
\usepackage{amsfonts}       
\usepackage{nicefrac}       
\usepackage{microtype}      
\usepackage{xcolor}         
\usepackage{graphicx}
\usepackage{subfig}
\usepackage{float}
\usepackage{wrapfig}
\usepackage[breakable, theorems, skins]{tcolorbox}
\usepackage{multirow}
\usepackage{makecell}
\usepackage{algorithm}
\usepackage{algorithmic}
\usepackage{setspace}
\usepackage[frozencache,cachedir=.]{minted}
\newtheorem{obs}{\textbf{Observation}}

\DeclareRobustCommand{\mybox}[2][gray!20]{%
\begin{tcolorbox}[   
        breakable,
        left=0pt,
        right=0pt,
        top=0pt,
        bottom=0pt,
        colback=#1,
        colframe=#1,
        width=\dimexpr\textwidth\relax, 
        enlarge left by=0mm,
        boxsep=2pt,
        arc=0pt,outer arc=0pt,
        ]
        #2
\end{tcolorbox}
}

\title{Dealing with Synthetic Data Contamination in Online Continual Learning}

\author{Maorong Wang$^{1}$ \quad Nicolas Michel$^{1,2}$ \quad Jiafeng Mao$^{1}$ \quad Toshihiko Yamasaki$^{1}$ \vspace{0.3em} \\
{\normalsize $^1$The University of Tokyo} \quad
{\normalsize $^2$Univ Gustave Eiffel, CNRS, LIGM} \quad \\
{\normalsize \texttt{\{ma\_wang, yamasaki\}@cvm.t.u-tokyo.ac.jp}} \quad \\
{\normalsize \texttt{nicolas.michel@univ-eiffel.fr}} \quad \\
{\normalsize \texttt{mao@hal.t.u-tokyo.ac.jp}} \quad \\
}

\begin{document}

\maketitle

\begin{abstract}



Image generation has shown remarkable results in generating high-fidelity realistic images, in particular with the advancement of diffusion-based models. However, the prevalence of AI-generated images may have side effects for the machine learning community that are not clearly identified. Meanwhile, the success of deep learning in computer vision is driven by the massive dataset collected on the Internet. The extensive quantity of synthetic data being added to the Internet would become an obstacle for future researchers to collect ``clean'' datasets without AI-generated content. Prior research has shown that using datasets contaminated by synthetic images may result in performance degradation when used for training. In this paper, we investigate the potential impact of contaminated datasets on Online Continual Learning (CL) research. We experimentally show that contaminated datasets might hinder the training of existing online CL methods. Also, we propose \textbf{E}ntropy \textbf{S}election with \textbf{R}eal-synthetic similarity \textbf{M}aximization (ESRM), a method to alleviate the performance deterioration caused by synthetic images when training online CL models. Experiments show that our method can significantly alleviate performance deterioration, especially when the contamination is severe. For reproducibility, the source code of our work is available at \url{https://github.com/maorong-wang/ESRM}.

\end{abstract}

\section{Introduction}
\label{sec:intro}
\vspace{-0.25cm}

Continual Learning (CL)~\cite{chen2018lifelong, parisi2019continual, wang2023comprehensive, de2021continual} solves the problem of learning from a sequence of ever-emerging machine-learning tasks without forgetting previously learned knowledge. Defined by learning manners, CL can be classified into two categories~\cite{aljundi2019gradient}: \textit{offline} CL and \textit{online} CL. In offline CL (\textit{i.e.} conventional CL), the learners can access the training dataset on \textbf{current} task multiple times before proceeding to the next task. In online CL, the training data also comes in a continual data stream, and the continual learner only sees the training data once. Besides learning manners, there are also three typical CL settings~\cite{van2019three}: Task-Incremental Learning (TIL), Domain-Incremental Learning (DIL), and Class-Incremental Learning (CIL). In this paper, we investigate the more challenging CIL setting in the online CL manner. 


Image generation with deep generative models has shown remarkable success. Thanks to denoising diffusion models~\cite{DDPM, StableDiffusion}, Internet users are capable of generating high-fidelity and realistic images within several seconds. Despite the astonishing quality of those images to human eyes, research has shown that AI-generated content may be harmful when used to train machine learning models, leading to potential performance deterioration~\cite{hataya2023will, martinez2023towards}, bias amplification~\cite{chen2024would}, loss of diversity~\cite{martinez2023towards}, etc. 

Recently, it has become a trend for researchers to collect datasets from the Internet, and synthetic data contamination would become a potential threat to the CL community. Moreover, the online CL is particularly affected as assessing the soundness of data in the online scenario is impractical. In this work, we first aim to investigate how this new form of dataset contamination might affect the existing Online CL methods. Then, we empirically observe the characteristics of synthetic data when used to train online CL models and form four observations of synthetic data properties in online CL, which might be of interest to the community. 
Moreover, we investigate synthetic data properties and exhibit specific differences in terms of entropy and representations when compared to real data. Guided by these properties, we propose \textbf{E}ntropy \textbf{S}election with \textbf{R}eal-synthetic similarity \textbf{M}aximization (ESRM), a method to alleviate the performance degradation caused by the synthetic contamination.
As a replay-based method, ESRM consists of two key components: Entropy Selection (ES) and Real-synthetic similarity Maximization (RM).
ES selects more realistic samples in the memory buffer to alleviate catastrophic forgetting. RM is a contrastive learning based optimization strategy, that aims to alleviate the performance artifact caused by synthetic data contamination. 

The major contribution of this paper can be summarized as follows: 

\begin{itemize}
    \item We investigate the potential impact of synthetic data contamination on existing online CL methods and outline four observations regarding the properties of synthetic data in continual scenarios.

    \item We propose ESRM, a method to alleviate the performance deterioration caused by synthetic data contamination.  

    \item Comprehensive experiments show that ESRM can successfully mitigate the performance deterioration caused by synthetic contamination, especially when the contamination is severe.
\end{itemize}
\section{Related Work}
\vspace{-0.25cm}

\paragraph{Synthetic data contamination.}
Recently, diffusion models~\cite{DDPM, StableDiffusion} have achieved high-fidelity image generation and surpassed GANs in terms of image quality and diversity. Likewise, text-to-image generation based on diffusion models can generate astonishing images that faithfully follow the users' text instructions. Furthermore, these generative models demonstrate excellent extrapolation capabilities (\textit{i.e.}, meaningfully combining concepts that would be nearly impossible to combine in reality), such as ``a photo of an astronaut riding a house''. Various generative models are open-sourced to the public, and users can use these models to generate realistic images in seconds.
However, while people are appreciating the new format of art and flooding the Internet with fabulous images, such synthetic images are difficult to differentiate from the real ones. Therefore, these generated images are becoming a potential source of contamination for the future datasets collected from the Internet.
Research has proven that synthetic data contamination may lead to a significant performance drop when supervising machine learning models in non-continual scenarios~\cite{hataya2023will, martinez2023towards}. Also, training deep models with such a contaminated dataset may give rise to bias amplification~\cite{chen2024would} and loss of diversity~\cite{martinez2023towards}. 
To tackle the issue caused by the synthetic contamination, researchers have proposed different strategies to detect the synthetic data with deep learning based detectors~\cite{UniFD,FatFormer}. However, the challenge brought by synthetic data contamination to the CL community is exclusive, and it is even more problematic in the online scenario since it is almost impossible to assess the quality of the training data, due to the unique challenge brought by the online setting.

\paragraph{Continual Learning.}
The mainstream CL strategies can be classified into four categories: regularization-based, parameter-isolation-based, prompt-based, and replay-based. Regularization-based methods~\cite{chaudhry2018riemannian, lee2017overcoming, aljundi2018memory, kirkpatrick2017overcoming, zenke2017continual} design and apply extra regularization terms to balance the learning and forgetting of CL learners. Parameter-isolation-based methods~\cite{fernando2017pathnet, rusu2016progressive, serra2018overcoming, rosenfeld2018incremental, aljundi2017expert} tackle the CL problem by allocating task-specific parameters. Prompt-based methods~\cite{L2P, Dualprompt} take the idea of prompt learning and use prompt pools against catastrophic forgetting. Replay-based methods~\cite{ER, DER, ERACE, OCM, GSA, OnPro} store a small portion of historical data with a memory buffer. 
Among all of the strategies, replay-based methods have prevailed in Online CL with better performance and simplicity. Early work~\cite{ER} proposed \textbf{Experience Replay (ER)}, suggesting using a random replay buffer to alleviate catastrophic forgetting. \textbf{Dark Experience Replay (DER++)}~\cite{DER} proposes to store the logits in the memory buffer and leverage the stored logits as dark knowledge to extend ER. \textbf{ER-ACE}~\cite{ERACE} is a variant of ER with asymmetric cross-entropy loss. \textbf{OCM}~\cite{OCM} alleviates catastrophic forgetting by maximizing the mutual information between current and past representations. \textbf{GSA}~\cite{GSA} addresses the cross-task class discrimination problem with gradient self-adaption. \textbf{OnPro}~\cite{OnPro} solves the shortcut learning problem with online prototype learning. In this paper, these methods are used as baselines to assess the impact of synthetic data contamination.

\begin{figure}[t]
  \centering
   \includegraphics[width=\linewidth]{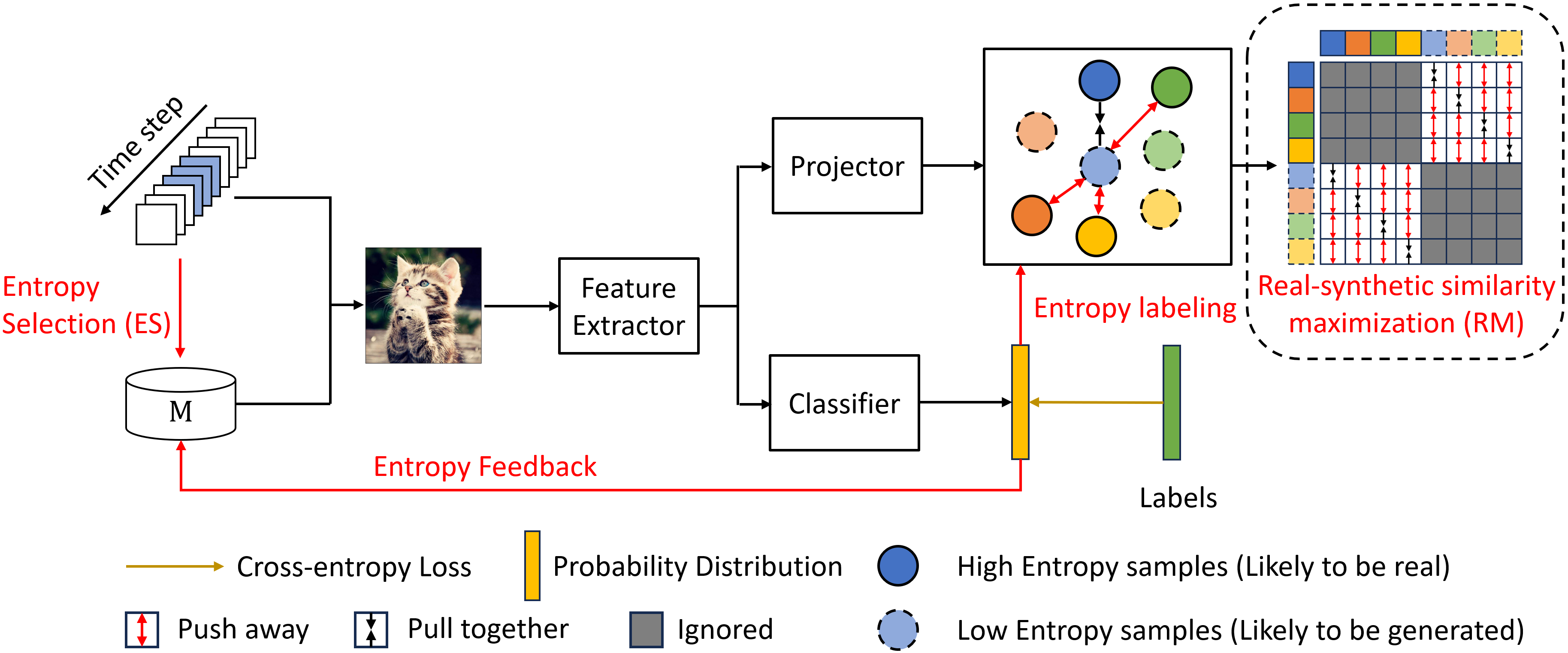}
   \caption{Overview of proposed ESRM framework for online CL. The proposed ESRM framework has two main components: Entropy Selection (ES) and Real-synthetic similarity Maximization (RM). Motivated by Obs.~\ref{obs:4} and Obs.~\ref{obs:2}, ES is a buffer management strategy designed to use entropy as a criterion to select more real samples in the memory buffer, thereby alleviating catastrophic forgetting and performance degradation caused by the contamination. RM aims to bridge the embedding gap between synthetic and real data, as noted in Obs.~\ref{obs:3}, using a contrastive learning technique.}
   \label{fig:teaser}
   \vspace{-10pt}
\end{figure}

\section{Preliminary}
\label{sec:preliminary}
\vspace{-0.2cm}

\subsection{Synthetic dataset generation}
\label{subsec:syn_gen}
\vspace{-0.1cm}
To simulate a dataset contaminated with synthetic data, we employed five diffusion-based models to generate synthetic counterparts of the original datasets. These twin datasets contain the same number of images and the same classes, while all images are synthetic. The models used in generation include Stable Diffusion XL~\cite{StableDiffusion}, Stable Diffusion v1.4, Stable Diffusion v2.1, VQDM~\cite{VQDM}, and GLIDE~\cite{GLIDE}. The synthetic data contamination was simulated across four benchmark datasets used in online CL, including CIFAR-10~\cite{CIFAR}, CIFAR-100~\cite{CIFAR}, TinyImageNet~\cite{Tiny}, and ImageNet-100~\cite{Imagenet, In100}. 

The generation of synthetic twin datasets is guided by the category names of the original dataset. For each class, we devised a simple yet effective prompt. For instance, for the class ``helicopter'', we employed the prompt ``an image of a helicopter''. After the generation by the diffusion model, we adjusted the image size to match that of the original dataset. 

In the experiments, we design two different settings:
\textbf{a)} all the synthetic twin datasets are generated from Stable Diffusion XL, one of the state-of-the-art generative models; and
\textbf{b)} Synthetic images are generated with the aforementioned five diffusion models, with each model contributing 20\% to the synthetic dataset.

For setting \textbf{a)}, we denote the generated dataset as SDXL-C10, SDXL-C100, SDXL-Tiny, and SDXL-In100, respectively, while for setting \textbf{b)}, we denote the generated dataset as Mix-C10, Mix-C100, and Mix-Tiny. Some examples of generated images and more detailed information about the synthetic twin dataset generation can be found in Appendix~\ref{apdx:generation}. Notably, the generation results in Fig.~\ref{fig:samples} reveal the lack of diversity for synthetic data compared with the real data.

\subsection{Simulation of synthetic data contamination}
\vspace{-0.1cm}
\label{subsec:sim}
To simulate the contamination by the synthetic data, we substitute a portion $P$ of the original dataset with its synthetic twin, where $P$ is the contamination ratio. We designate these contaminated datasets using specific notations. For example, we denote the CIFAR-100 contaminated by SDXL-C100 as C100/SDXL. More details about the simulation of contamination are included in Appendix~\ref{apdx:simulation}.
In the following sections, we will investigate the effect of synthetic contamination with these simulated datasets.

\vspace{-0.25cm}
\section{Synthetic Data Contamination in Online CL}
\label{sec:syn}
\vspace{-0.25cm}
In this section, we explore the potential impact of synthetic data contamination on the existing online CL methods and exhibit specific properties of synthetic data. 

\subsection{Contamination as a cause of performance degradation}
\vspace{-0.1cm}
When dealing with synthetic data contamination, it is crucial to measure the impact of such contamination on existing methods. In that sense, we train ResNet-18 models in the online continual setting on the contaminated dataset C10/SDXL, C100/SDXL, Tiny/SDXL, and In-100/SDXL with representative online CL methods. We observed that as the contamination ratio $P$ increases, the performance of all existing methods drops significantly, as shown in Table~\ref{tab:main}.  
Detailed information about the experiment settings can be found in Sec.~\ref{sec:exp}. With such experiments, we can form the following observation:

\mybox{
\begin{obs}
\label{obs:1}
As a source of potential contamination, synthetic data is harmful to the performance of existing online CL methods. Performance degradation increases as contamination becomes more severe.
\end{obs}
}

\begin{figure}
\centering
\begin{minipage}{.54\textwidth}
  \centering
  \subfloat[ER]{
       \includegraphics[height=3.3cm]{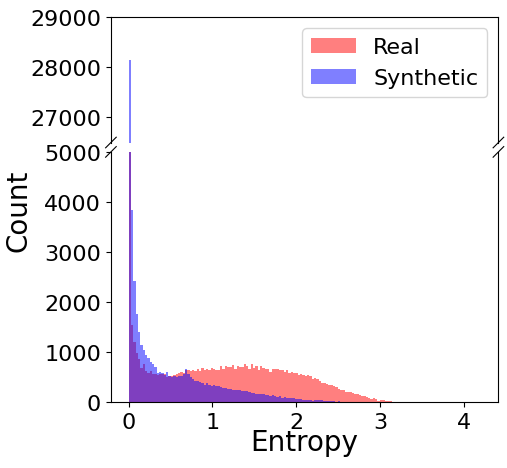}}
    \label{1a}
    \subfloat[OnPro]{
        \includegraphics[height=3.3cm]{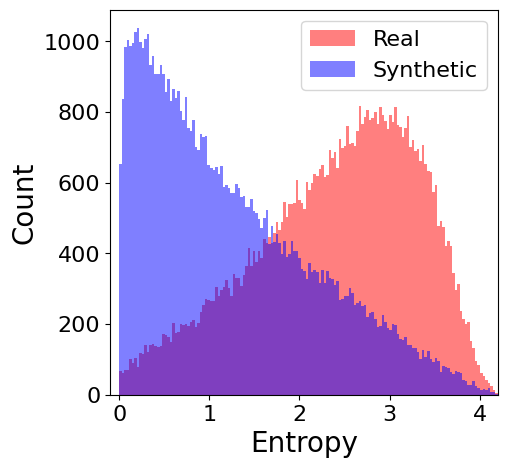}}
    \label{1b}
  \captionof{figure}{The entropy distribution of the training dataset produced by ER and OnPro on In-100/SDXL ($P=50\%$) at the end of the training.}
  \label{fig:ent}
\end{minipage}%
\hfill
\begin{minipage}{.43\textwidth}
  \centering
  \includegraphics[width=0.8\linewidth]{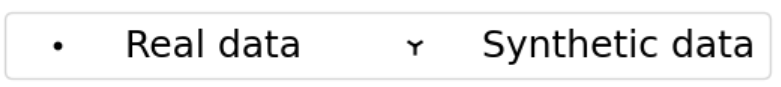}
  \subfloat[ER]{
       \includegraphics[height=2.85cm]{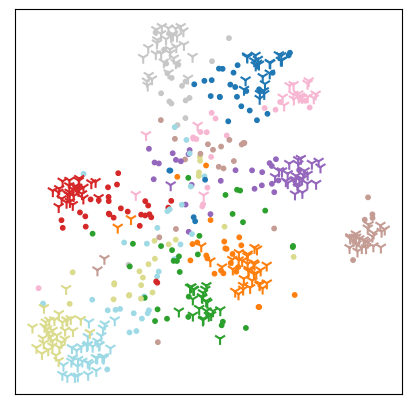}}
    \label{2a}
    \subfloat[OnPro]{
        \includegraphics[height=2.85cm]{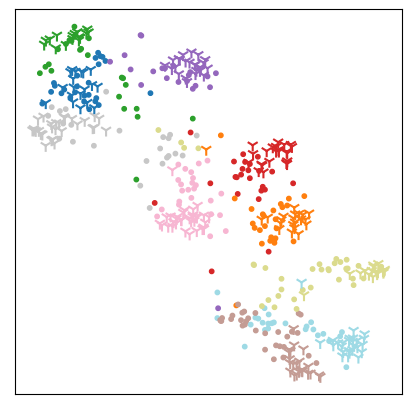}}
    \label{2b}
  \captionof{figure}{T-SNE visualization of the memory data at the end of training on In-100/SDXL ($P=50\%$). For clarity, only the first 10 classes are visualized.}
  \label{fig:tsne_baseline}
\end{minipage}
\vspace{-15pt}
\end{figure}

\subsection{Detecting synthetic data matters}
\vspace{-0.1cm}

\begin{wraptable}{r}{.5\linewidth}
\centering
\vspace{-10pt}
\resizebox{\linewidth}{!}{\begin{tabular}{cccc}
\toprule
Method & Memory strategy & \makecell{$P = 70\%$ \\ Acc. $\uparrow$} & \makecell{$P = 80\%$ \\ Acc. $\uparrow$} \\
\midrule
\multirow{3}{*}{ER} & Real Only & 38.44{\scriptsize ±0.90} & 38.13{\scriptsize ±1.34}\\
& Random & 32.82{\scriptsize ±1.62} &	31.33{\scriptsize ±1.33}\\
& Synthetic Only & 22.45{\scriptsize ±1.88} &	22.01{\scriptsize ±1.35}\\
\bottomrule
\end{tabular}}
\caption{The performance of ER with different memory strategies on C100/SDXL dataset, with different contamination ratio $P$.}\label{tab:mem}
\end{wraptable} 

Replay-based methods are characterized by the existence of a replay buffer, which helps alleviate forgetting and implicitly improves network plasticity~\cite{CCLDC}. The quality of data in the replay buffer intuitively influences network performance. Motivated by Obs.~\ref{obs:1}, we hypothesize that the presence of synthetic data in the replay buffer might degrade performance. We trained ER on the C100/SDXL dataset, with extra information on the samples' synthetic status (\textit{i.e.}, whether the image is real or synthetic). We employed two memory strategies: storing only real data in the replay buffer and storing only synthetic data. The results showed in Table~\ref{tab:mem} indicate that knowing the synthetic status and storing only real data in the replay buffer can achieve performance on par with the no-contamination scenario, even at high contamination ratios. For instance, at a contamination ratio of $P=80\%$, ER achieves an accuracy of 38.13\% on C100/SDXL when only real images are stored in the replay buffer. This result is comparable to the accuracy achieved when training on the clean CIFAR-100 dataset (38.70\%), which contains four times more real data. These findings also demonstrate the potential of synthetic models to enhance performance through data generation and augmentation in an online continual learning scenario.

\mybox{
\begin{obs}
\label{obs:4}
The memory buffer plays a key role in replay-based methods, and storing real samples in the memory buffer is effective against performance degradation caused by contamination.
\end{obs}
}

\subsection{Synthetic data properties}
\textbf{Lower entropy distribution.} One noteworthy characteristic of synthetic data is its lower entropy distribution compared to real data, as observed from the perspective of continual learners. Fig.~\ref{fig:ent} shows the entropy distribution produced by a representative method ER and a state-of-the-art method OnPro when trained on the contaminated dataset. The values in the histogram are calculated at the end of the training on the whole training dataset (In-100/SDXL, $P = 50\%$). From the figure, we can spot a salient distribution difference between synthetic data and real data. Moreover, the synthetic samples have a peak in entropy distribution close to 0. Extra entropy histogram with other baselines can be found in Appendix~\ref{subsec:addition_ent}. Thus, we conclude with another finding:

\mybox{
\begin{obs}
\label{obs:2}
Compared with real data, synthetic data entropy tends to be lower.
\end{obs}
}


\textbf{Feature gap in the embedding space.} We find another intriguing property of synthetic data in the feature embeddings. Fig.~\ref{fig:tsne_baseline} shows the t-SNE~\cite{tsne} visualization of the memory data produced by ER and OnPro on the In-100/SDXL dataset. For clarity, we only visualize the embeddings of the first 10 classes. With the synthetic data contamination, the features of the synthetic samples are better clustered than the real data. The pattern of the clustering of synthetic data indicates the ease of classification, which is on par with the limited diversity of synthetic data (cf. Fig.~\ref{fig:samples}), and the low entropy distribution (cf. Obs.~\ref{obs:2}). Moreover, the embeddings of real data are inferior and fail to align with the superior embeddings of synthetic samples, which explains the performance degradation of inference on real test datasets. Extra visualization produced by other baselines is illustrated in Appendix~\ref{subsec:addition_tsne}.

\mybox{
\begin{obs}
\label{obs:3}
With the limited diversity of synthetic data, the synthetic data are better clustered than the real data, leading to a misalignment in the embedding space between synthetic samples and real samples. Such misalignment likely contributes to performance deterioration.
\end{obs}
}

\vspace{-0.25cm}
\section{Proposed Method}
\vspace{-0.25cm}


Fig.~\ref{fig:teaser} presents the main framework of our proposed ESRM to alleviate the performance degradation caused by synthetic data contamination. In this section, we introduce the two components of ESRM: Entropy Selection (ES) and Real-synthetic similarity Maximization (RM). Then, we explain the whole ESRM framework.

\vspace{-0.1cm}
\subsection{Entropy selection}
\vspace{-0.1cm}
\begin{wrapfigure}{r}{0.4\textwidth}
    \centering
    \vspace{-12pt}
    \includegraphics[width=1.0\linewidth]{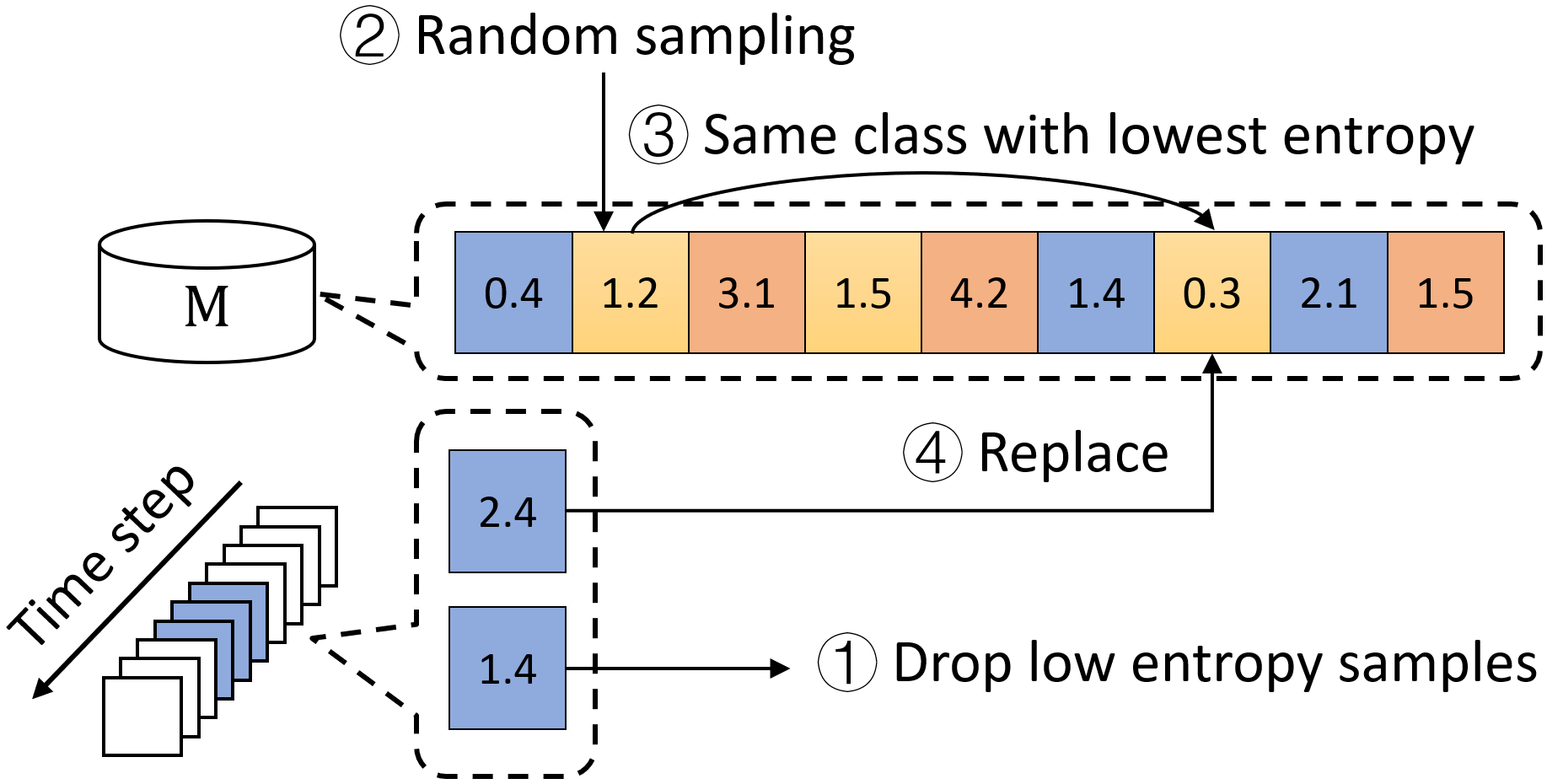}
    \caption{Overview of the proposed Entropy Selection strategy. The color of the samples indicates the class, and the number in the samples represents the entropy predicted by the learner.}
    \label{fig:es}
    \vspace{-10pt}
\end{wrapfigure}

The introduced ES aims to select more representative samples in the memory buffer. For replay-based methods, having high-quality samples in the memory buffer helps alleviate forgetting and achieve better overall performance. 
As per Obs.~\ref{obs:4}, selecting real images into the memory buffer can provide more representative and reliable features and therefore lead to better performance. Motivated by this, we propose ES, a memory management strategy. Guided by Obs.~\ref{obs:2} (synthetic data has lower entropy distributions), the core idea of ES is to select more real samples in the memory buffer
based on the entropy distribution of the current batch.

At the beginning of the training, ES initializes an empty buffer. When a new batch comes, ES drops 50\% of the batch samples with lower entropy, and stores the remaining samples, along with their entropy of the prediction. Once the buffer is full, as shown in Fig.~\ref{fig:es}, ES takes four steps to replace the elements in the buffer. Firstly, ES drops 50\% of low entropy samples in the incoming batch. Then, ES uses Reservoir Sampling~\cite{isele2018selective, vitter1985random} to decide whether to keep or discard the remaining incoming samples. If Reservoir Sampling decides to keep the incoming sample, it will nominate a buffer sample. After that, ES checks the class of the nominated sample and chooses a sample of the same class with the lowest entropy in the memory buffer. And finally, ES replaces the chosen memory sample with the incoming sample, along with its entropy. Moreover, all entropy values in the memory buffer are updated by the current model's prediction at the end of each task. The pseudo-code of ES is given in Appendix~\ref{sec:pseudocode}.


\vspace{-0.1cm}
\subsection{Real-synthetic similarity maximization}
\vspace{-0.1cm}

Prior to introducing the loss function of RM, we present the network structures of our ESRM. As shown in Fig.~\ref{fig:teaser}, the continual learner consists of three components: a feature extractor $f$, a projection head $g$, and a classifier $\phi$. The output dimension of the projection head $g$ is set to 128. For each sample $x$ from incoming data stream $X^{new}$, the projected embedding $z$ can be formulated as: 
\begin{equation}
    z=g(f(x)).
\end{equation}
As per Obs.~\ref{obs:3}, the gap in the embedding space might be a cause for the performance degradation in the contamination setting. To tackle this issue, we propose RM. The main idea of RM is to maximize the cosine similarity between the features of real and synthetic data. Motivated by the seminal supervised contrastive loss~\cite{SupCon}, we propose our variant RM loss.



RM aims to maximize the similarity between two groups of data $X_1$ and $X_2$. We define the loss to match the similarity of samples in group $X_1$ to group $X_2$ as: 
\begin{equation}
    \mathcal{L}_{M}(X_1, X_2)= \sum_{i \in I_1} \frac{-1}{|P(i)|} \sum_{p \in P(i)} log \frac{exp(z_i \cdot z_p / \tau)}{\sum_{d \in I_2} exp(z_i \cdot z_d / \tau)},
\end{equation}
where $I_1=\{i : x_i \in X_1\}, I_2=\{d : x_d \in X_2\}$ are the set of the indices of $X_1, X_2$, respectively. And $P(i) = \{p \in X_2 : y_p = y_i\}$ is the set of the indices of positive samples in group $X_2$, which share the same class with $x_i$. $\tau$ is the temperature hyperparameter which is set to 0.07. Since $\mathcal{L}_{M}(X_1, X_2)$ only optimize samples in the group $X_1$, in the optimization, it is used together with $\mathcal{L}_{M}(X_2, X_1)$. Due to RM aims to maximize the inter-group similarity between the $X_1$ group and the $X_2$ group, we do not need to perform augmentations as shown in similar work, such as SimCLR~\cite{SimCLR} and SupCon~\cite{SupCon}. Intuitively, the way to handle the similarity matrix is illustrated in Fig.\ref{fig:teaser}.

For an incoming batch $X^{new}$, we use entropy criteria to split it into two groups $X^{new}_+$ and $X^{new}_-$ of the same size. Group $X^{new}_+$ includes 50\% of the samples with the highest entropy, and as per Obs.~\ref{obs:2}, tend to contain more real images. On the contrary, group $X^{new}_-$ contains low entropy samples which tend to be synthetic. To alleviate the gap in feature embeddings mentioned in Obs.~\ref{obs:3}, we maximize the inter-group similarity between $X^{new}_+$ and $X^{new}_-$ with $ \mathcal{L}_{M}(X^{new}_+, X^{new}_-)$ and $\mathcal{L}_{M}(X^{new}_-, X^{new}_+)$. Moreover, to align the holistic feature embeddings between the stream data $X^{new}$ and memory data $X^{mem}$, we also applied $\mathcal{L}_{M}(X^{new}, X^{mem})$ and $\mathcal{L}_{M}(X^{mem}, X^{new})$ in the loss function.

Thus, the proposed RM to alleviate the feature gap in Obs.~\ref{obs:3} can be achieved by employing $\mathcal{L}_{RM}$:
\begin{equation}
    \mathcal{L}_{RM} = \mathcal{L}_{M}(X^{new}_+, X^{new}_-) + \mathcal{L}_{M}(X^{new}_-, X^{new}_+) + \mathcal{L}_{M}(X^{new}, X^{mem}) + \mathcal{L}_{M}(X^{mem}, X^{new}).
\end{equation}
\subsection{Overall framework of ESRM}

The overall framework of ESRM is shown in Fig.~\ref{fig:teaser}. Besides ES and RM, following~\cite{CCLDC}, ESRM employs a self-distillation technique to alleviate the overconfidence problem of the replay-based methods. For the combined batch $X = (X^{new}, X^{mem})$, we apply the self-distillation as:
\begin{equation}
    \mathcal{L}_{SDC} = D_{KL}(\phi(f(X))/t, \overline{\phi(f(aug(X)))}/t),
\end{equation}
where $D_{KL}(\cdot)$ is the Kullback-Leibler divergence, $t$ is another temperature hyperparameter which is set to 4, $\overline{\phi(f(aug(X)))}$ is the fixed copy of $\phi(f(aug(X)))$, without gradient propagation, and $aug(\cdot)$ is the data augmentation used in the training process, with detailed information in Appendix~\ref{apdx:aug}. 

Thus, the total loss of ESRM can be formulated as:
\begin{equation}
    \label{eq:all}
    \mathcal{L}_{ESRM} = \mathcal{L}_{CE} + \lambda_1\mathcal{L}_{SDC} + \lambda_2 \mathcal{L}_{RM},
\end{equation}
where $\mathcal{L}_{CE} = CE(\phi(f(X)), y) + CE(\phi(f(aug(X))), y)$ is the cross-entropy loss, and $\lambda_1, \lambda_2$ are the balancing hyperparamteres. We set $\lambda_1 = 1$ and $ \lambda_2 = 0.5$ after a small hyperparameter search as illustrated in the Appendix~\ref{apdx:hpsearch}. 

\begin{table*}
    \centering
    \resizebox{\textwidth}{!}{
    \begin{tabular}{lcccccc}
\toprule
\multicolumn{1}{c}{Dataset} & CIFAR10 & \multicolumn{5}{c}{C10/SDXL}\\
\cmidrule(lr){1-1}
\cmidrule(lr){2-2}
\cmidrule(lr){3-7}

\multicolumn{1}{c}{Ratio $P$} & 0\% & 50\% & 70\% & 80\% & 90\% & 95\% \\
\midrule
ER &		63.93{\scriptsize ±2.40} &	60.40{\scriptsize ±1.54}(-3.53) &	57.07{\scriptsize ±3.64}(-6.86) &	54.18{\scriptsize ±3.42}(-9.75) &	50.69{\scriptsize ±3.56}(-13.24) &	47.39{\scriptsize ±2.76}(-16.54) \\
DER++ &	64.31{\scriptsize ±2.63} &	60.24{\scriptsize ±2.02}(-4.07) &	56.11{\scriptsize ±2.99}(-8.20) &	51.62{\scriptsize ±3.28}(-12.69) &	44.43{\scriptsize ±3.00}(-19.88) &	40.46{\scriptsize ±2.92}(-23.85) \\
ERACE &	60.19{\scriptsize ±2.51} &	56.17{\scriptsize ±2.08}(-4.02) &	50.70{\scriptsize ±2.66}(-9.49) &	46.86{\scriptsize ±4.61}(-13.33) &	41.86{\scriptsize ±2.40}(-18.33) &	37.56{\scriptsize ±2.46}(-22.63) \\
OCM &		72.66{\scriptsize ±1.61} &	69.70{\scriptsize ±1.52}(-2.96) &	66.68{\scriptsize ±1.69}(-5.98) &	63.79{\scriptsize ±1.57}(-8.87) &	60.41{\scriptsize ±1.36}(-12.25) &	57.07{\scriptsize ±1.42}(-15.59) \\
GSA &		66.91{\scriptsize ±1.57} &	63.47{\scriptsize ±1.89}(-3.44) &	59.44{\scriptsize ±2.33}(-7.47) &	56.60{\scriptsize ±2.63}(-10.31) &	49.93{\scriptsize ±2.97}(-16.98) &	45.77{\scriptsize ±2.60}(-21.14) \\
OnPro &	\textbf{74.87{\scriptsize ±1.58}} &	\textbf{72.46{\scriptsize ±1.36}}(-2.41) &  \textbf{68.79{\scriptsize ±1.17}}(-6.08) &	66.07{\scriptsize ±1.23}(-8.80) &	62.37{\scriptsize ±1.70}(-12.50) &	56.41{\scriptsize ±2.59}(-18.46) \\
ESRM &	67.35{\scriptsize ±1.14} &	67.95{\scriptsize ±0.88}\textbf{(0.6)} &	67.47{\scriptsize ±1.43}\textbf{(0.12)} &	\textbf{66.81{\scriptsize ±1.59}(-0.54)} &	\textbf{64.59{\scriptsize ±1.59}(-2.76)} &	\textbf{60.04{\scriptsize ±2.29}(-7.31)} \\

\midrule
\multicolumn{1}{c}{Dataset} & CIFAR100 & \multicolumn{5}{c}{C100/SDXL}\\
\cmidrule(lr){1-1}
\cmidrule(lr){2-2}
\cmidrule(lr){3-7}

\multicolumn{1}{c}{Ratio $P$} & 0\% & 50\% & 70\% & 80\% & 90\% & 95\% \\
\midrule
ER 	&	38.70{\scriptsize ±1.45} &	36.37{\scriptsize ±1.39}(-2.33) &	32.82{\scriptsize ±1.62}(-5.88) &	31.33{\scriptsize ±1.33}(-7.37) &	27.09{\scriptsize ±1.02}(-11.61) &	25.56{\scriptsize ±1.23}(-13.14) \\
DER++ &	37.62{\scriptsize ±2.30} &	32.35{\scriptsize ±2.72}(-5.27) &	28.32{\scriptsize ±2.45}(-9.30) &	25.57{\scriptsize ±2.44}(-12.05) &	19.56{\scriptsize ±1.88}(-18.06) &	16.00{\scriptsize ±1.64}(-21.62) \\
ERACE &	39.82{\scriptsize ±1.37} &	34.15{\scriptsize ±1.33}(-5.67) &	28.16{\scriptsize ±1.37}(-11.66) &	25.13{\scriptsize ±1.30}(-14.69) &	19.37{\scriptsize ±1.49}(-20.45) &	16.14{\scriptsize ±1.29}(-23.68) \\
OCM 	&	42.01{\scriptsize ±1.07} &	40.21{\scriptsize ±1.15}(-1.80) &	37.54{\scriptsize ±1.36}(-4.47) &	34.78{\scriptsize ±1.32}(-7.23) &	31.40{\scriptsize ±1.51}(-10.61) &	28.84{\scriptsize ±1.28}(-13.17) \\
GSA 	&	42.27{\scriptsize ±1.53} &	39.21{\scriptsize ±1.13}(-3.06) &	35.21{\scriptsize ±1.62}(-7.06) &	32.64{\scriptsize ±1.84}(-9.63) &	27.62{\scriptsize ±1.15}(-14.65) &	23.88{\scriptsize ±1.61}(-18.39) \\
OnPro &	41.47{\scriptsize ±1.09} &	39.26{\scriptsize ±0.72}(-2.21) & 35.64{\scriptsize ±0.60}(-5.83) &	33.20{\scriptsize ±0.70}(-8.27) &	30.20{\scriptsize ±0.84}(-11.27) &	26.77{\scriptsize ±1.19}(-14.70) \\
ESRM 	&	\textbf{47.72{\scriptsize ±0.87}} &	\textbf{46.57{\scriptsize ±0.92}(-1.15)} &	\textbf{45.92{\scriptsize ±0.42}(-1.80)} &	\textbf{44.48{\scriptsize ±0.41}(-3.24)} &	\textbf{40.99{\scriptsize ±0.70}(-6.73)} &	\textbf{37.45{\scriptsize ±0.56}(-10.27)} \\
\midrule
\multicolumn{1}{c}{Dataset} & Tiny & \multicolumn{5}{c}{Tiny/SDXL}\\
\cmidrule(lr){1-1}
\cmidrule(lr){2-2}
\cmidrule(lr){3-7}

\multicolumn{1}{c}{Ratio $P$} & 0\% & 50\% & 70\% & 80\% & 90\% & 95\% \\
\midrule
ER & 25.06{\scriptsize ±1.81} &	18.03{\scriptsize ±1.69}(-7.03) &	13.59{\scriptsize ±1.86}(-11.47) &	11.29{\scriptsize ±1.46}(-13.77) &	6.38{\scriptsize ±0.89}(-18.68) &	3.88{\scriptsize ±0.65}(-21.18) \\
DER++ & 19.40{\scriptsize ±3.71} &	12.55{\scriptsize ±2.26}(-6.85) &	9.71{\scriptsize ±1.41}(-9.69) &	7.46{\scriptsize ±1.45}(-11.94) &	4.49{\scriptsize ±0.83}(-14.91) &	2.81{\scriptsize ±0.41}\textbf{(-16.59)} \\
ERACE & 26.38{\scriptsize ±1.03} &	17.04{\scriptsize ±0.88}(-9.34) &	11.23{\scriptsize ±0.69}(-15.15) &	7.83{\scriptsize ±1.16}(-18.55) &	4.09{\scriptsize ±0.58}(-22.29) &	2.54{\scriptsize ±0.55}(-23.84) \\
OCM & 31.94{\scriptsize ±1.44} &	25.21{\scriptsize ±0.65}(-6.73) &	20.14{\scriptsize ±1.16}(-11.80) &	16.16{\scriptsize ±0.64}(-15.78) &	10.35{\scriptsize ±0.58}(-21.59) &	5.37{\scriptsize ±0.63}(-26.57) \\
GSA & 25.34{\scriptsize ±1.01} &	15.59{\scriptsize ±2.29}(-9.75) &	12.55{\scriptsize ±1.65}(-12.79) &	9.31{\scriptsize ±1.20}(-16.03) &	5.95{\scriptsize ±0.83}(-19.39) &	3.87{\scriptsize ±0.55}(-21.47) \\
OnPro & 26.38{\scriptsize ±2.18} &	16.92{\scriptsize ±1.22}(-9.46) &	13.23{\scriptsize ±1.26}(-13.15) &	8.82{\scriptsize ±1.26}(-17.56) &	4.80{\scriptsize ±1.12}(-21.58) &	2.68{\scriptsize ±0.58}(-23.70) \\
ESRM & \textbf{32.15{\scriptsize ±1.20}} &	\textbf{29.36{\scriptsize ±0.53}(-2.79)} &	\textbf{27.81{\scriptsize ±1.02}(-4.34)} &	\textbf{25.7{\scriptsize ±1.16}(-6.45)} &	\textbf{19.09{\scriptsize ±1.57}(-13.06)} &	\textbf{13.02{\scriptsize ±0.79}}(-19.13) \\

\midrule
\multicolumn{1}{c}{Dataset} & In-100 & \multicolumn{5}{c}{In-100/SDXL}\\
\cmidrule(lr){1-1}
\cmidrule(lr){2-2}
\cmidrule(lr){3-7}
\multicolumn{1}{c}{Ratio $P$} & 0\% & 50\% & 70\% & 80\% & 90\% & 95\% \\
\midrule

ER 	&	33.35{\scriptsize ±1.84} &	30.49{\scriptsize ±0.91}(-2.81) &	26.47{\scriptsize ±0.64}(-6.83) &	23.61{\scriptsize ±0.68}(-9.69) &	21.09{\scriptsize ±0.61}(-12.21) &	18.54{\scriptsize ±0.92}(-14.76) \\
DER++ &	34.89{\scriptsize ±2.27} &	29.98{\scriptsize ±4.35}(-4.77) &	26.84{\scriptsize ±1.72}(-7.91) &	23.72{\scriptsize ±2.10}(-11.03) &	20.57{\scriptsize ±1.67}(-14.18) &	17.52{\scriptsize ±2.06}(-17.23) \\
ERACE &	38.43{\scriptsize ±1.45} &	32.96{\scriptsize ±1.55}(-5.41) &	28.99{\scriptsize ±0.73}(-9.38) &	25.28{\scriptsize ±0.77}(-13.09) &	20.51{\scriptsize ±1.13}(-17.86) &	16.91{\scriptsize ±0.64}(-21.46) \\
OCM 	&	26.70{\scriptsize ±2.36} &	26.43{\scriptsize ±0.53}(-0.27) &	23.70{\scriptsize ±1.11}(-3.00) &	23.61{\scriptsize ±2.00}(-3.09) &	22.21{\scriptsize ±1.09}(-4.49) &	20.56{\scriptsize ±1.20}\textbf{(-6.14)} \\
GSA 	&	\textbf{40.85{\scriptsize ±1.04}} &	37.35{\scriptsize ±1.23}(-3.68) &	32.34{\scriptsize ±1.30}(-8.69) &	29.47{\scriptsize ±0.44}(-11.56) &	25.15{\scriptsize ±0.74}(-15.88) &	21.89{\scriptsize ±0.25}(-19.14) \\
OnPro &	38.47{\scriptsize ±1.13} &	36.17{\scriptsize ±2.70}(-2.58) &	34.04{\scriptsize ±2.51}(-4.71) &	33.01{\scriptsize ±1.07}(-5.74) &	29.34{\scriptsize ±2.00}(-9.41) &	25.14{\scriptsize ±1.26}(-13.61) \\
ESRM	&	39.72{\scriptsize ±1.05} &	\textbf{39.64{\scriptsize ±0.76}(-0.08)} &	\textbf{39.62{\scriptsize ±0.90}(-0.10)} &	\textbf{39.53{\scriptsize ±0.83}(-0.19)} &	\textbf{36.58{\scriptsize ±0.80}(-3.14)} &	\textbf{32.86{\scriptsize ±0.99}}(-6.86) \\

\bottomrule
\end{tabular}

    }
    \caption{Average Accuracy (\%; higher is better) on four benchmark datasets with different contamination ratios $P$. Numbers in parentheses indicate the performance degradation due to contamination compared to the clean setting. The average and deviation over five runs are reported for ImageNet-100 and 10 runs for other datasets.}
    \label{tab:main}
    \vspace{-8pt}
\end{table*}

\vspace{-0.25cm}
\section{Experiments}
\vspace{-0.2cm}
\label{sec:exp}

\subsection{Experiment setup}
\vspace{-0.1cm}

\textbf{Datasets.} In the experiments, we used four benchmark datasets in evaluation, including CIFAR-10/100, TinyImageNet, and ImageNet-100. All of the datasets are split into tasks containing non-overlapping classes. The details about the task split are available in Appendix~\ref{apdx:dataset}.


\textbf{Baselines.}
We evaluate the effectiveness of ESRM against six representative and state-of-the-art baselines, including ER~\cite{ER}, DER++~\cite{DER}, ERACE~\cite{ERACE}, OCM~\cite{OCM}, GSA~\cite{GSA}, and OnPro~\cite{OnPro}.

\textbf{Implementation details.}
We use full-width ResNet-18 (not pre-trained) as the backbone for all experiments. For a fair comparison, we conduct a hyperparameter search on CIFAR-100 (Memory Size = 5K) and apply the same hyperparameter to all settings. Stream batch size is set to 10 and memory batch size is set to 64. We do not use multiple updates trick for incoming batches as detailed in~\cite{aljundi2019online}. Detailed information about task allocation, hyperparameter search protocol, and data augmentation can be found in Appendix~\ref{apdx:implementation}.

\textbf{Buffer size.} For CIFAR-10 and CIFAR-100 experiments, the buffer size is set to 1,000 and 5,000, respectively. For the harder TinyImageNet experiments, the buffer size is set to 10,000. The buffer size of ImageNet-100 is set to 5,000. Appendix~\ref{subsec:mem} demonstrates more experiments with different memory buffer sizes.

\begin{table*}[t]
    \centering
    \resizebox{.7\textwidth}{!}{
    \begin{tabular}{lcccccc}
\toprule
\multicolumn{1}{c}{Dataset} & In-100 & \multicolumn{5}{c}{In-100/SDXL}\\
\cmidrule(lr){1-1}
\cmidrule(lr){2-2}
\cmidrule(lr){3-7}

\multicolumn{1}{c}{Ratio $P$} & 0\% & 50\% & 70\% & 80\% & 90\% & 95\% \\
\midrule

ER 	&	53.87{\scriptsize ±1.27}&	52.12{\scriptsize ±1.03}&	48.38{\scriptsize ±0.79}&	45.21{\scriptsize ±1.24}&	39.24{\scriptsize ±1.58}&	35.27{\scriptsize ±0.93}\\
DER++ &	61.14{\scriptsize ±2.38}&	56.99{\scriptsize ±1.85}&	53.56{\scriptsize ±2.86}&	50.26{\scriptsize ±2.06}&	44.08{\scriptsize ±1.89}&	37.92{\scriptsize ±2.19}\\
ERACE &	49.25{\scriptsize ±1.94}&	43.83{\scriptsize ±0.76}&	37.92{\scriptsize ±1.49}&	33.51{\scriptsize ±0.85}&	26.78{\scriptsize ±1.14}&	22.70{\scriptsize ±0.75}\\
OCM 	&	22.78{\scriptsize ±1.62}&	19.37{\scriptsize ±0.28}&	19.81{\scriptsize ±0.53}&	19.43{\scriptsize ±0.77}&	20.44{\scriptsize ±0.25}&	18.98{\scriptsize ±1.44}\\
GSA 	&	61.83{\scriptsize ±1.72}&	58.15{\scriptsize ±1.99}&	53.63{\scriptsize ±0.46}&	49.55{\scriptsize ±1.25}&	42.35{\scriptsize ±0.38}&	36.16{\scriptsize ±0.90}\\
OnPro &	39.28{\scriptsize ±0.84}&	38.98{\scriptsize ±3.16}&	40.07{\scriptsize ±1.72}&	38.45{\scriptsize ±2.80}&	36.45{\scriptsize ±1.51}&	32.94{\scriptsize ±1.33}\\
ESRM &	\textbf{74.85{\scriptsize ±0.61}}&	\textbf{71.25{\scriptsize ±0.74}}&	\textbf{66.51{\scriptsize ±0.39}}&	\textbf{62.16{\scriptsize ±0.39}}&	\textbf{55.30{\scriptsize ±0.50}}&	\textbf{50.41{\scriptsize ±0.97}}\\

\bottomrule
\end{tabular}

    }
    \caption{Learning Accuracy (\%; higher is better) on In-100/SDXL with various contamination ratio $P$. }
    \label{tab:la}

    \centering
    \resizebox{.7\textwidth}{!}{
    \begin{tabular}{lcccccc}
\toprule
\multicolumn{1}{c}{Dataset} & In-100 & \multicolumn{5}{c}{In-100/SDXL}\\
\cmidrule(lr){1-1}
\cmidrule(lr){2-2}
\cmidrule(lr){3-7}

\multicolumn{1}{c}{Ratio $P$} & 0\% & 50\% & 70\% & 80\% & 90\% & 95\% \\
\midrule

ER 	&	39.38{\scriptsize ±3.51}&	42.20{\scriptsize ±1.70}&	45.95{\scriptsize ±1.52}&	48.38{\scriptsize ±2.10}&	47.12{\scriptsize ±1.74}&	47.83{\scriptsize ±2.26}\\
DER++ &	42.26{\scriptsize ±5.72}&	48.35{\scriptsize ±8.20}&	51.00{\scriptsize ±5.01}&	53.73{\scriptsize ±4.78}&	54.28{\scriptsize ±5.20}&	54.93{\scriptsize ±6.77}\\
ERACE &	24.21{\scriptsize ±2.17}&	25.22{\scriptsize ±3.29}&	24.23{\scriptsize ±2.29}&	26.10{\scriptsize ±3.97}&	26.26{\scriptsize ±4.28}&	28.80{\scriptsize ±3.92}\\
OCM 	&	\textbf{4.67{\scriptsize ±1.66}}&	\textbf{4.38{\scriptsize ±2.16}}&	\textbf{5.96{\scriptsize ±3.12}}&	\textbf{4.86{\scriptsize ±1.62}}&	\textbf{6.45{\scriptsize ±1.78}}&	\textbf{6.52{\scriptsize ±1.31}}\\
GSA 	&	35.51{\scriptsize ±2.12}&	37.50{\scriptsize ±1.72}&	40.79{\scriptsize ±3.08}&	41.72{\scriptsize ±0.94}&	41.83{\scriptsize ±2.04}&	40.55{\scriptsize ±0.56}\\
OnPro &	15.96{\scriptsize ±2.23}&	23.12{\scriptsize ±4.69}&	22.87{\scriptsize ±3.14}&	23.89{\scriptsize ±1.70}&	24.72{\scriptsize ±1.80}&	28.17{\scriptsize ±2.93}\\
ESRM &	49.79{\scriptsize ±1.61}&	46.02{\scriptsize ±1.19}&	40.83{\scriptsize ±1.29}&	36.90{\scriptsize ±1.74}&	34.61{\scriptsize ±1.82}&	35.40{\scriptsize ±2.38}\\

\bottomrule
\end{tabular}

    }
    \caption{Relative Forgetting (\%; lower is better) on In-100/SDXL with various contamination ratio $P$. }
    \label{tab:rf}
\vspace{2pt}
\centering
    \begin{minipage}[t]{0.47\linewidth}\centering
    \resizebox{0.85\width}{!}{\begin{tabular}{ccc}
\toprule
Memory strategy & \makecell{$P = 70\%$ \\ Acc. $\uparrow$} & \makecell{$P = 80\%$ \\ Acc. $\uparrow$} \\
\midrule
Real Only (Idealized) & 47.43{\scriptsize ±0.62} &	46.85{\scriptsize ±0.76}  \\
ES (Ours) & 45.92{\scriptsize ±0.42} &	44.48{\scriptsize ±0.41} \\
Random & 44.84{\scriptsize ±0.80} &	42.62{\scriptsize ±0.87}\\
Synthetic Only (Worst) & 25.91{\scriptsize ±0.79} &	24.74{\scriptsize ±0.91}\\
\bottomrule
\end{tabular}}
    \caption{Comparison of different memory strategies on C100/SDXL dataset, with different contamination ratio $P$. }\label{tab:ablation_es}
    \end{minipage}
\hfill%
    \begin{minipage}[t]{0.47\linewidth}\centering
    \resizebox{0.85\width}{!}{\begin{tabular}{cccc}
\toprule
Method & \makecell{$P = 70\%$ \\ Acc. $\uparrow$} & \makecell{$P = 80\%$ \\ Acc. $\uparrow$} \\
\midrule
Baseline & 42.61{\scriptsize ±0.84} & 41.22{\scriptsize ±0.65}\\
w/o $\mathcal{L}_{SDC}$ &  44.30{\scriptsize ±0.78} &	42.59{\scriptsize ±0.69}  \\
w/o $\mathcal{L}_{RM}$ & 44.03{\scriptsize ±0.73} & 42.32{\scriptsize ±0.52} \\
ESRM & 45.92{\scriptsize ±0.42} &	44.48{\scriptsize ±0.41} \\
\bottomrule
\end{tabular}

}
    \caption{Ablation of loss functions on C100/SDXL dataset, with different contamination ratio $P$. ``Baseline'' indicates ES+$\mathcal{L}_{CE}$. }\label{tab:ablation_rm}
    \end{minipage}
    \vspace{-12pt}
\end{table*}

\subsection{Results and analysis}
\label{subsec:results}
\textbf{Final Average Accuracy.} Table~\ref{tab:main} shows the final average accuracy of learners trained with four datasets, including C10/SDXL, C100/SDXL, Tiny/SDXL, and In-100/SDXL, with different contamination ratios $P$. More results on C10/Mix, C100/Mix, and Tiny/Mix are given in Appendix~\ref{subsec:mix_perform}. For the six baselines, we can notice a significant performance drop when synthetic data contamination appears. Notably, when the contamination is severe (contamination ratio $P \ge 70\%$), the performance degradation is significant. Also, it can be observed that ESRM is less impacted by synthetic data contamination. For most datasets and contamination ratio $P$, the ESRM performance drop is the lowest of the compared methods. This is remarkably true for large values of $P$.

Apart from the robustness against synthetic data contamination, the absolute performance of ESRM is also attractive. In most cases, ESRM outperforms the baseline methods by a large margin. More interestingly, for some datasets, such as CIFAR-100 and ImageNet-100, even with an extreme contamination ratio, ESRM can still achieve a substantial performance, while the performance of most baseline methods is unsatisfactory.

\textbf{Plasticity and Stability Metrics.}
We measure the model's plasticity and stability with Learning Accuracy (LA)~\cite{LA} and Relative Forgetting (RF)~\cite{CCLDC}, respectively. As shown in Table~\ref{tab:la} and~\ref{tab:rf}, for baseline methods, both plasticity and stability performance are hindered by synthetic data contamination. From the model plasticity perspective, ESRM alleviates the problem with a larger plasticity. For the model stability, ESRM solves the problem of stability degradation with the presence of contamination. It is notable that with RM loss and self-distillation loss, ESRM implicitly trades off some model stability in favor of plasticity. Plasticity and stability performance on other datasets are available in Appendix~\ref{subsec:plasticityext}.

\textbf{Domain-Incremental Learning Results.} While Class-Incremental Learning (CIL) setting is the standard evaluation protocol in online CL, we also evaluated the performance of ESRM in Domain-Incremental Learning (DIL) scenarios. We conducted the experiment with the 20 coarse labels of the CIFAR-100 dataset. Since the 100 classes in CIFAR-100 are grouped into 20 superclasses with 5 fine-grained classes for each superclass, we split the CIFAR-100 dataset with 5 domain increment steps. For each step, we feed the model with the training data of a fine-grained class from each superclass. Because the model only classifies 20 coarse labels, we refer to this dataset as DIL-CIFAR20. Similar to the simulated CIFAR100/SDXL dataset, we replace the images in the DIL-CIFAR20 dataset with its Stable Diffusion XL generated counterpart with a contamination ratio P, as per the protocol in Sec.~\ref{subsec:sim}.

Table~\ref{tab:dil} shows the final average accuracy with different contamination ratios. Notably, we adapted the CIL-specifically designed components in OnPro and GSA to the DIL scenario, and the performance suffered a decent loss. We did not report ERACE results because its Asymmetric Cross Entropy (ACE) loss converges to standard cross-entropy loss in the DIL scenario, making it equivalent to vanilla ER.
The experimental results show that ESRM can yield robust performance against domain shift in the DIL setting, under different synthetic contamination situations, which validates the efficiency of ESRM under DIL settings.

\subsection{Ablation studies}

\textbf{Effect of ES.} 
To evaluate the effect of ES, we substitute ES with three different memory strategies: random sampling, storing real data only, and storing synthetic data only. Note that storing real or synthetic data requires knowing the ground truth of an image's synthetic status, which is not practical in realistic settings. Additionally, storing only real data is an idealized case for memory management, while storing only synthetic data represents the worst-case scenario. As shown in Table~\ref{tab:ablation_es}, both ES and random reservoir sampling~\cite{isele2018selective, vitter1985random} outperform the worst-case scenario by a large margin. Moreover, ES outperforms the random sampling significantly.

\textbf{Effects of loss terms.}
We also conduct experiments to verify the effects of loss terms in Eq.~\ref{eq:all}. As shown in Table~\ref{tab:ablation_rm}, Both $\mathcal{L}_{SDC}$ and $\mathcal{L}_{RM}$ can benefit the final average accuracy of the classification. Furthermore, the combination of the two loss terms can further improve the final accuracy, validating that both terms complement each other. 

\begin{table*}
    \centering
    \resizebox{.8\textwidth}{!}{
        
\begin{tabular}{lcccc}
\toprule
\multicolumn{1}{c}{Dataset} & DIL-CIFAR20 & \multicolumn{3}{c}{DIL-C20/SDXL}\\
\cmidrule(lr){1-1}
\cmidrule(lr){2-2}
\cmidrule(lr){3-5}
\multicolumn{1}{c}{Ratio $P$} & 0\% & 50\% & 70\%  & 80\%\\
\midrule
ER&		53.56{\scriptsize ±1.42} &	51.91{\scriptsize ±1.63}(-1.65) &	49.61{\scriptsize ±0.76}(-3.95) &	47.25{\scriptsize ±0.73}(-6.31)\\
DER++&	56.43{\scriptsize ±0.65} &	52.46{\scriptsize ±1.19}(-3.97) &	48.73{\scriptsize ±1.50}(-7.70) &	44.60{\scriptsize ±1.79}(-11.83)\\
OCM&		56.02{\scriptsize ±1.17} &	53.87{\scriptsize ±0.57}(-2.15) &	52.69{\scriptsize ±0.93}(-3.33) &	50.57{\scriptsize ±0.80}(-5.45)\\
GSA&		53.67{\scriptsize ±2.60} &	51.54{\scriptsize ±1.76}(-2.13) &	47.16{\scriptsize ±1.76}(-6.51) &	45.55{\scriptsize ±1.17}(-8.12)\\
OnPro&	31.81{\scriptsize ±1.21} &	30.43{\scriptsize ±0.72}(-1.38) &	29.19{\scriptsize ±0.82}(-2.62) &	27.79{\scriptsize ±0.86}\textbf{(-4.02)}\\
Ours&		\textbf{64.27{\scriptsize ±0.46}} &	\textbf{63.35{\scriptsize ±0.70}(-0.92)} &	\textbf{61.86{\scriptsize ±0.54}(-2.41)} &	\textbf{59.94{\scriptsize ±0.71}}(-4.33)\\
\bottomrule
\end{tabular}
    }
    \caption{Final Average Accuracy (\%; higher is better) on DIL-C20/SDXL dataset. Numbers in parentheses indicate the performance degradation due to synthetic contamination compared to the clean setting. The average and deviation over 10 runs are reported.}
    \label{tab:dil}
\end{table*}
\begin{figure}
\centering

\begin{minipage}{.45\textwidth}
    \includegraphics[width=0.8\linewidth]{Figs/tsnelegend.png}
  \centering
    \includegraphics[width=.6\linewidth]{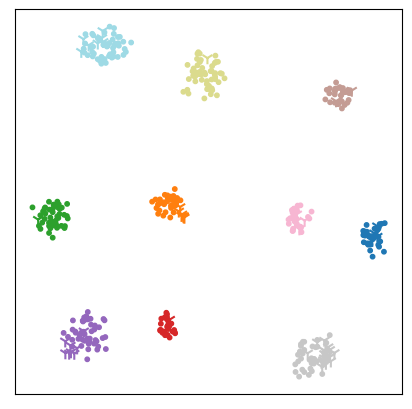}
    \captionof{figure}{T-SNE visualization of memory data produced by ESRM at the end of training on the In-100/SDXL ($P=50\%$) dataset. For clarity, only the first 10 classes are visualized.}
    \label{fig:tsne_ersm}

\end{minipage}%
\hfill
\begin{minipage}{.5\textwidth}
  \centering
    \includegraphics[width=.9\linewidth]{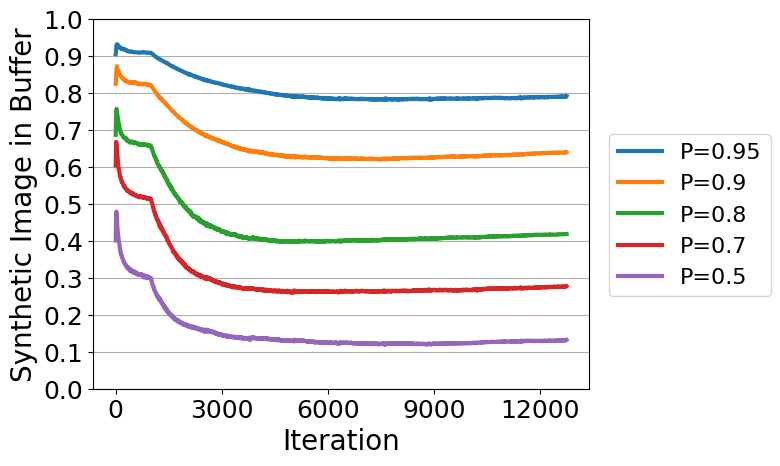}
      \captionof{figure}{The percentage of synthetic data in the memory buffer throughout the training of ESRM on the In-100/SDXL dataset with different contamination ratios ($P$). The average value of 5 runs is plotted.}
      \label{fig:mem_dyn}

\end{minipage}
\vspace{-.3cm}

\end{figure}

\section{Discussions}
\vspace{-0.25cm}

\begin{wrapfigure}{r}{0.3\textwidth}
    \centering
    \vspace{-12pt}
      \includegraphics[width=\linewidth]{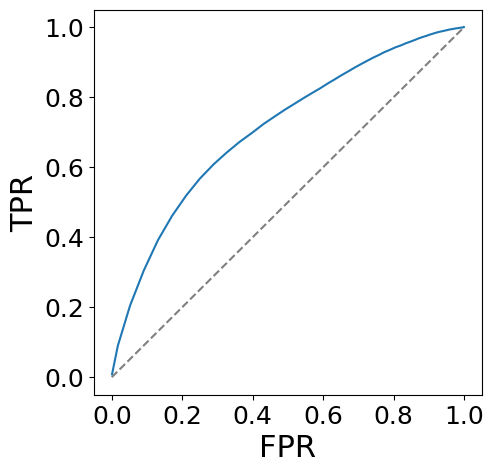}
      \caption{The ROC curve of the model trained with ESRM on the In-100/SDXL dataset ($P=50\%$) in predicting the synthetic status of samples in the training dataset. Real samples are regarded as positives and synthetic samples as negatives.}
      \label{fig:ROC}
    \vspace{-10pt}
\end{wrapfigure}

\textbf{The alleviation of feature misalignment.} 
As mentioned in Obs.~\ref{obs:3}, baseline methods suffer performance degradation due to the misalignment between the inferior feature embedding of real images and the superior feature embedding of synthetic samples. Fig.~\ref{fig:tsne_ersm} presents the t-SNE visualization of memory data at the end of training of ESRM on the In-100/SDXL dataset. Similar to Fig.~\ref{fig:tsne_baseline}, only the first 10 classes are visualized for clarity.  Compared to ER and OnPro, the embeddings of synthetic and real samples in ESRM are better aligned, facilitated by the RM.

\textbf{Training dynamics of memory buffer.} 
  To intuitively demonstrate the effect of ES, we visualized the percentage of synthetic data in the memory buffer throughout the whole training process. Fig.~\ref{fig:mem_dyn} displays the curve of the percentage of synthetic data in the memory buffer when the model is trained with ESRM on In-100/SDXL with different contamination ratios $P$. To generate this curve, we checked the memory buffer every 10 iterations. As shown in the figure, the percentage of synthetic data is close to the contamination ratio $P$ in the early stages of training. As training progresses, the amount of synthetic data decreases. This trend intuitively illustrates the effect of ES in selecting real samples.
Furthermore, we take a model trained with ESRM at the end of training and use the model's entropy as the criterion to categorize the synthetic status of samples in the training dataset and generate an ROC curve, as shown in Fig.~\ref{fig:ROC}. We regard real data as positive and use models trained with ESRM on In-100/SDXL ($P=50\%$) to categorize the samples in the training set. The AUC of the ROC curve is 0.7098, showing the effect of the entropy criterion in discriminating real and synthetic samples.

\vspace{-0.3cm}
\section{Conclusion}
\vspace{-0.25cm}
With the widespread availability of advanced generative models, the prevalence of AI-generated images appears inevitable, posing a potential challenge for researchers attempting to collect datasets devoid of AI-generated content from the Internet. In this paper, we examine the potential side effects of AI-powered image generation on the continual learning community. First, we experimentally demonstrate that synthetic data has become a potential source of data pollution. We spot a catastrophic performance loss when the contaminated datasets are used to train continual learning models. Based on our experiments, we identify and summarize four typical characteristics of synthetic data when involved in the training of continual learners. Additionally, we propose ESRM, a method designed to alleviate performance deterioration, maintaining satisfactory performance even with highly contaminated datasets. Lastly, we hope our work highlights the need for improved regulation and systematic control over generated data, such as watermarking AI-generated content before publication. Internet data is a valuable resource accumulated over decades. We believe ensuring the integrity of Internet data is crucial for the future soundness of AI development.

\bibliographystyle{plain}
\bibliography{mainbib}

\newpage
\appendix

\vspace{-.3cm}
\section{Limitations}
\vspace{-.3cm}
The paper investigates the potential impact of synthetic data contamination on online CL research and proposes a method to alleviate side effects caused by the contamination. Nevertheless, our research has some limitations. Firstly, in the evaluation, we only use five generative models, including Stable Diffusion v1.4, Stable Diffusion v2.1, Stable Diffusion XL, VQDM, and GLIDE. Other excellent commercial generative works, such as Midjourney and DALL-E, are not included in the data generation. With limited computation/resources, we could not exhaust all generative methods.

Secondly, the way we produce the generated dataset is simple: we use prompts like ``an image of a <class\_name>'' in the generation. In reality, the users' prompts are usually more diverse. Some works use LLMs to simulate more realistic prompts, where we leave a further in-depth analysis for future research.

\begin{spacing}{0.8}
\begin{algorithm}[t]
\footnotesize
\begin{minted}{python}
# model: continual learning model.
# criteria(): loss function as in Eq. 5.
# n_seen_so_far: count of images seen by the buffer.
# random(): function returns random values in (0,1).
# buffer_labels: 1D Tensor of size [buffer_size] storing the labels.
# buffer_ent: 1D Tensor of size [buffer_size] storing the entropy values.
# update_ent(): Update all entropy values in the buffer.

n_seen_so_far = 0
for t in tasks:
    for img, label in dataloader:
        # train the network with stream data
        model.train()
        mem_img, mem_label = mem_sample()
        c_img, c_label = concat((img, mem_img), (label, mem_label)) # combined batch
        loss = criteria(model(aug(c_img)), c_label) # Eq. 5.
        loss.backward()
        optimizer.step()
        
        # ES updates
        # calculate the entropy criteria of stream data
        model.eval()
        logits = model(img)
        prob = softmax(logits)
        entropy = -torch.sum(prob * torch.log(prob))

        # update buffer
        threshold = torch.quantile(entropy, 0.5) # Step 1 in ES
        stream_img = img[entropy > threshold]
        stream_label = label[entropy > threshold]
        stream_entropy = entropy[entropy > threshold]

        for x, y, ent in zip(stream_img, stream_label, stream_ent):
            nominate = int(random()*(n_seen_so_far + 1)) # Step 2 in ES
            if n_seen_so_far < buffer_size: # if the buffer is not full
                nominate = n_seen_so_far
                replace_data(nominate, x, y, ent)
                n_seen_so_far += 1
            elif nominate < buffer_size:
                nominate_class = buffer_labels[nominate] # Step 3 in ES
                idx = buffer_ent[buffer_labels == nominate_class].argmin()
                replace_data(idx, x, y, ent) # Step 4 in ES
                n_seen_so_far += 1

    update_ent()
            
\end{minted}
\vspace{-10pt}
\caption{PyTorch-like pseudo-code of ES.}
\label{code:pseudo_code}
\end{algorithm}
\end{spacing}

\vspace{-.3cm}
\section{Pseudo code for ES}
\label{sec:pseudocode}
\vspace{-.25cm}

The PyTorch-like pseudo code showing how the ES updates the memory buffer is shown in Alg.~\ref{code:pseudo_code}.

\begin{figure*}[t]
    \subfloat[ER]{
       \includegraphics[width=0.24\linewidth]{Figs/Ent_ER.png}}
    \hfill
    \subfloat[DER++]{
       \includegraphics[width=0.24\linewidth]{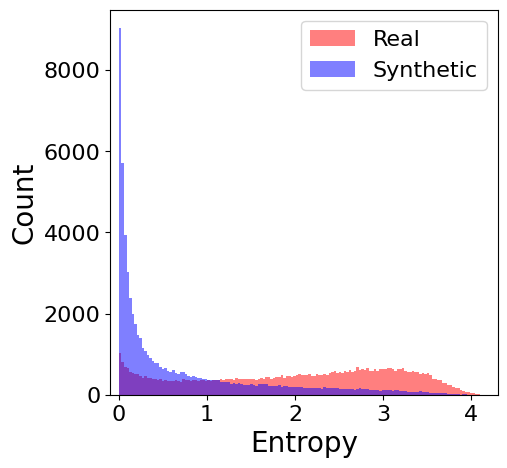}}
    \hfill
    \subfloat[ERACE]{
       \includegraphics[width=0.24\linewidth]{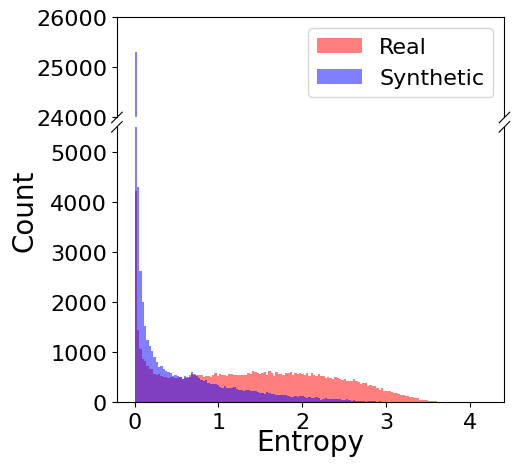}}
    \hfill
    \subfloat[OCM]{
       \includegraphics[width=0.24\linewidth]{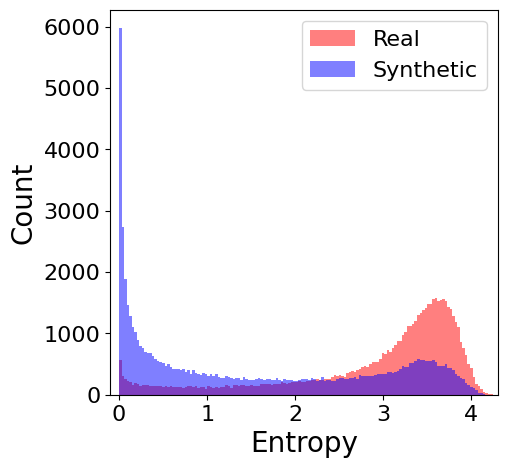}}
    \\
    \hspace*{4em}
    \subfloat[GSA]{
       \includegraphics[width=0.24\linewidth]{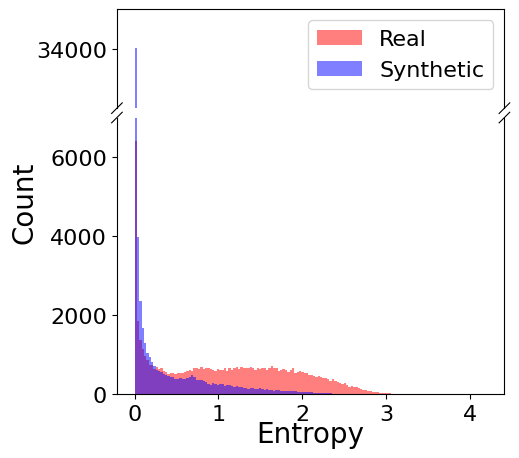}}
    \hfill
    \subfloat[OnPro]{
       \includegraphics[width=0.24\linewidth]{Figs/Ent_OnPro.png}}
    \hfill
    \subfloat[ESRM]{
       \includegraphics[width=0.24\linewidth]{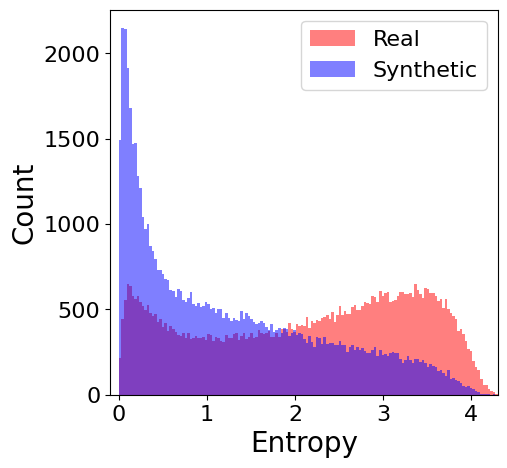}}
    \hspace*{4em}
   \caption{The entropy distribution of the training set produced by all methods on In-100/SDXL $P=50\%$ at the end of the training.}
   \label{fig:ent_extra}
\end{figure*}

\begin{figure*}[t]
\centering
    \includegraphics[width=0.3\linewidth]{Figs/tsnelegend.png}\\
    \subfloat[ER]{
       \includegraphics[width=0.24\linewidth]{Figs/tsne_er.png}}
    \hfill
    \subfloat[DER++]{
       \includegraphics[width=0.24\linewidth]{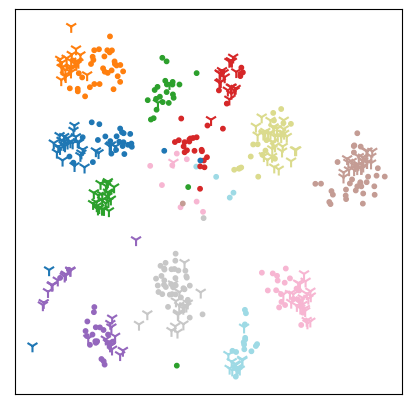}}
    \hfill
    \subfloat[ERACE]{
       \includegraphics[width=0.24\linewidth]{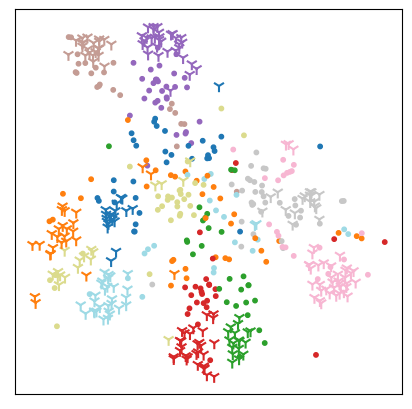}}
    \hfill
    \subfloat[OCM]{
       \includegraphics[width=0.24\linewidth]{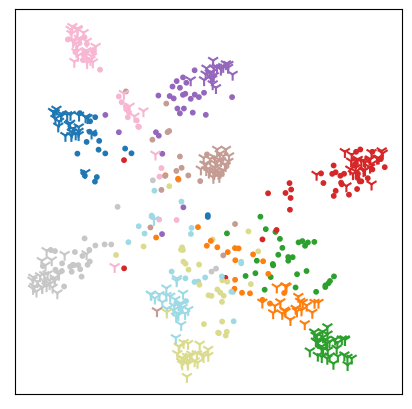}}
    \\
    \vspace{-1em}
    \hspace*{4em}
    \subfloat[GSA]{
       \includegraphics[width=0.24\linewidth]{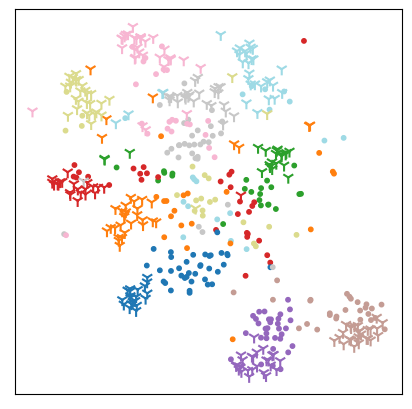}}
    \hfill
    \subfloat[OnPro]{
       \includegraphics[width=0.24\linewidth]{Figs/tsne_onpro.png}}
    \hfill
    \subfloat[ESRM]{
       \includegraphics[width=0.24\linewidth]{Figs/tsne_esrm.png}}
    \hspace*{4em}
   \caption{T-SNE visualization of the memory data at the end of training on In-100/SDXL ($P=50\%$). For clarity, only the first 10 classes are visualized.}
   \label{fig:tsne_extra}
\end{figure*}

\section{Extra Experiments}

\subsection{Entropy distribution of other baselines}
\label{subsec:addition_ent}
As mentioned in Sec.~\ref{sec:syn}, we show the entropy distribution produced by other baselines, when trained on In-100/SDXL ($P=50\%$). Fig.~\ref{fig:ent_extra} illustrates the entropy distribution histogram, which is calculated at the end of the training on the whole contaminated dataset. Similar to the result in Fig.~\ref{fig:ent}, the entropy distribution of the synthetic data is saliently lower than the entropy distribution of the real data.

\subsection{T-SNE visualization of other baselines}
\label{subsec:addition_tsne}
We show additional experiments of t-SNE visualization of memory data produced by other baseline methods. As shown in Fig.~\ref{fig:tsne_extra}, we visualize the memory embeddings of different baselines on the In-100/SDXL ($P=50\%$) dataset. For clarity, we only visualize the first 10 classes. Similar to the results in Fig.~\ref{fig:tsne_baseline}, the synthetic data are better clustered compared with the real data. This proves the Obs.~\ref{obs:3}.

\subsection{Performance on C10/Mix, C100/Mix and Tiny/Mix Dataset}
\label{subsec:mix_perform}
As mentioned in Sec.~\ref{sec:exp}, we would like to include the experiment results on C10/Mix, C100/Mix, and Tiny/Mix datasets. The final average accuracy of different methods on the dataset is included in Table~\ref{tab:mix}. Similar to the results in Table~\ref{tab:main}, ESRM has less performance degradation and better performance in most cases.

\subsection{Plasticity and stability performance on other datasets}
\label{subsec:plasticityext}
As mentioned in Sec.~\ref{subsec:results}, the plasticity and stability metrics of methods on other datasets are demonstrated in Table~\ref{tab:laext} and~\ref{tab:rfext}. Similar to the results in Table~\ref{tab:la} and~\ref{tab:rf}, in most settings, the plasticity metric (LA) and stability metric (RF) of baseline methods drop with an increased contamination ratio $P$. For the model plasticity, ESRM alleviates the issue with a larger plasticity. From a stability perspective, ESRM addresses the issue of stability degradation with the presence of contamination.

\subsection{The impact of buffer size.}
\label{subsec:mem}
In Sec.~\ref{subsec:results}, we evaluate the effectiveness of ESRM in a limited buffer size setting. To evaluate our methods' scalability against different buffer sizes $M$ and contamination ratio $P$, we compare the accuracy of different methods on C100/SDXL with different buffer sizes $M$, as shown in Table~\ref{tab:memc100}. Similar to the results in Table~\ref{tab:main}, ESRM can obtain better performance when the dataset is contaminated with synthetic data.

\begin{table*}
    \centering
    \resizebox{\textwidth}{!}{
    \begin{tabular}{lcccccc}
\toprule
\multicolumn{1}{c}{Dataset} & CIFAR10 & \multicolumn{5}{c}{C10/Mix}\\
\cmidrule(lr){1-1}
\cmidrule(lr){2-2}
\cmidrule(lr){3-7}

\multicolumn{1}{c}{Ratio $P$} & 0\% & 50\% & 70\% & 80\% & 90\% & 95\% \\
\midrule

ER 	&	63.93{\scriptsize ±2.40} &	61.74{\scriptsize ±2.36}(-2.19) &	57.54{\scriptsize ±2.04}(-6.39) &	57.18{\scriptsize ±2.51}(-6.75) &	53.71{\scriptsize ±2.32}(-10.22) &	52.21{\scriptsize ±2.92}(-11.72) \\
DER++ &	64.31{\scriptsize ±2.63} &	61.62{\scriptsize ±2.57}(-2.69) &	59.64{\scriptsize ±4.24}(-4.67) &	57.31{\scriptsize ±3.36}(-7.00) &	53.28{\scriptsize ±4.10}(-11.03) &	51.68{\scriptsize ±3.16}(-12.63) \\
ERACE &	60.19{\scriptsize ±2.51} &	58.21{\scriptsize ±2.38}(-1.98) &	55.01{\scriptsize ±2.59}(-5.18) &	51.15{\scriptsize ±3.45}(-9.04) &	49.16{\scriptsize ±3.10}(-11.03) &	47.74{\scriptsize ±2.45}(-12.45) \\
OCM 	&	72.66{\scriptsize ±1.61} &	70.13{\scriptsize ±1.38}(-2.53) &	66.62{\scriptsize ±1.69}(-6.04) &	65.08{\scriptsize ±1.63}(-7.58) &	62.86{\scriptsize ±1.39}(-9.80) &	61.60{\scriptsize ±1.99}(-11.06) \\
GSA 	&	66.91{\scriptsize ±1.57} &	63.05{\scriptsize ±2.02}(-3.86) &	62.45{\scriptsize ±1.52}(-4.46) &	60.94{\scriptsize ±1.50}(-5.97) &	57.51{\scriptsize ±2.58}(-9.40) &	55.84{\scriptsize ±1.43}(-11.07) \\
OnPro &	\textbf{74.87{\scriptsize ±1.58}} &	\textbf{72.45{\scriptsize ±1.97}}(-2.42) &	\textbf{69.79{\scriptsize ±1.78}}(-5.08) &	\textbf{67.13{\scriptsize ±1.41}}(-7.74) &	64.23{\scriptsize ±1.17}(-10.64) &	63.63{\scriptsize ±0.85}(-11.24) \\
ESRM 	&	67.35{\scriptsize ±1.14} &	68.04{\scriptsize ±1.33}\textbf{(0.69)} &	67.60{\scriptsize ±1.16}\textbf{(0.25)} &	66.73{\scriptsize ±1.03}\textbf{(-0.62)} &	\textbf{64.74{\scriptsize ±0.88}(-2.61)} &	\textbf{64.38{\scriptsize ±1.60}(-2.97)} \\

\midrule
\multicolumn{1}{c}{Dataset} & CIFAR100 & \multicolumn{5}{c}{C100/Mix}\\
\cmidrule(lr){1-1}
\cmidrule(lr){2-2}
\cmidrule(lr){3-7}

\multicolumn{1}{c}{Ratio $P$} & 0\% & 50\% & 70\% & 80\% & 90\% & 95\% \\
\midrule

ER 	&	38.70{\scriptsize ±1.45} &	36.54{\scriptsize ±1.40}(-2.16) &	33.82{\scriptsize ±1.13}(-4.88) &	32.33{\scriptsize ±0.77}(-6.37) &	30.24{\scriptsize ±0.80}(-8.46) &	28.36{\scriptsize ±1.22}(-10.34) \\
DER++ &	37.62{\scriptsize ±2.30} &	35.38{\scriptsize ±2.65}(-2.24) &	31.58{\scriptsize ±1.91}(-6.04) &	29.71{\scriptsize ±2.19}(-7.91) &	26.75{\scriptsize ±1.51}(-10.87) &	24.78{\scriptsize ±1.17}(-12.84) \\
ERACE &	39.82{\scriptsize ±1.37} &	35.65{\scriptsize ±1.04}(-4.17) &	31.44{\scriptsize ±2.28}(-8.38) &	29.21{\scriptsize ±1.30}(-10.61) &	26.24{\scriptsize ±1.17}(-13.58) &	23.42{\scriptsize ±1.41}(-16.4) \\
OCM 	&	42.01{\scriptsize ±1.07} &	39.78{\scriptsize ±0.94}(-2.23) &	37.19{\scriptsize ±1.19}(-4.82) &	35.51{\scriptsize ±1.58}(-6.50) &	33.37{\scriptsize ±1.04}(-8.64) &	32.01{\scriptsize ±0.63}(-10.00) \\
GSA 	&	42.27{\scriptsize ±1.53} &	39.88{\scriptsize ±1.54}(-2.39) &	36.72{\scriptsize ±0.68}(-5.55) &	34.27{\scriptsize ±1.59}(-8.00) &	31.83{\scriptsize ±1.42}(-10.44) &	30.17{\scriptsize ±1.61}(-12.10) \\
OnPro &	41.47{\scriptsize ±1.09} &	38.78{\scriptsize ±1.27}(-2.69) &	36.53{\scriptsize ±0.63}(-4.94) &	34.48{\scriptsize ±0.58}(-6.99) &	32.08{\scriptsize ±1.07}(-9.39) &	31.06{\scriptsize ±1.71}(-10.41) \\
ESRM 	&	\textbf{47.72{\scriptsize ±0.87}} &	\textbf{45.58{\scriptsize ±0.88}(-2.14)} &	\textbf{44.02{\scriptsize ±0.65}(-3.70)} &	\textbf{43.81{\scriptsize ±0.95}(-3.91)} &	\textbf{41.10{\scriptsize ±0.71}(-6.62)} &	\textbf{38.13{\scriptsize ±0.48}(-9.59)} \\

\midrule
\multicolumn{1}{c}{Dataset} & Tiny & \multicolumn{5}{c}{Tiny/Mix}\\
\cmidrule(lr){1-1}
\cmidrule(lr){2-2}
\cmidrule(lr){3-7}

\multicolumn{1}{c}{Ratio $P$} & 0\% & 50\% & 70\% & 80\% & 90\% & 95\% \\
\midrule

ER 	&	25.06{\scriptsize ±1.81} &	17.28{\scriptsize ±1.88}(-7.78) &	13.52{\scriptsize ±1.49}(-11.54) &	10.26{\scriptsize ±1.40}(-14.80) &	6.73{\scriptsize ±0.96}(-18.33) &	3.46{\scriptsize ±0.68}(-21.60) \\
DER++ &	19.40{\scriptsize ±3.71} &	13.86{\scriptsize ±2.44}(-5.54) &	10.56{\scriptsize ±0.72}(-8.84) &	7.35{\scriptsize ±1.78}(-12.05) &	3.66{\scriptsize ±0.61}(-15.74) &	1.59{\scriptsize ±0.46}\textbf{(-17.81)} \\
ERACE &	26.38{\scriptsize ±1.03} &	18.70{\scriptsize ±1.23}(-7.68) &	13.99{\scriptsize ±0.81}(-12.39) &	10.35{\scriptsize ±1.03}(-16.03) &	5.62{\scriptsize ±0.67}(-20.76) &	3.01{\scriptsize ±0.47}(-23.37) \\
OCM 	&	31.94{\scriptsize ±1.44} &	24.14{\scriptsize ±1.17}(-7.80) &	19.18{\scriptsize ±1.18}(-12.76) &	15.06{\scriptsize ±0.83}(-16.88) &	8.82{\scriptsize ±0.93}(-23.12) &	4.28{\scriptsize ±0.76}(-27.66) \\
GSA 	&	25.34{\scriptsize ±1.01} &	17.11{\scriptsize ±1.40}(-8.23) &	13.49{\scriptsize ±1.65}(-11.85) &	10.02{\scriptsize ±0.96}(-15.32) &	6.00{\scriptsize ±0.86}(-19.34) &	3.26{\scriptsize ±0.68}(-22.08) \\
OnPro &	26.38{\scriptsize ±2.18} &	17.03{\scriptsize ±1.67}(-9.35) &	11.84{\scriptsize ±1.42}(-14.54) &	8.82{\scriptsize ±1.24}(-17.56) &	3.79{\scriptsize ±0.61}(-22.59) &	1.72{\scriptsize ±0.62}(-24.66) \\
ESRM 	&	\textbf{32.15{\scriptsize ±1.20}} &	\textbf{27.37{\scriptsize ±1.24}(-4.78)} &	\textbf{25.65{\scriptsize ±1.12}(-6.50)} &	\textbf{23.52{\scriptsize ±1.24}(-8.63)} &	\textbf{17.29{\scriptsize ±1.16}(-14.86)} &	\textbf{11.50{\scriptsize ±0.74}}(-20.65) \\

\bottomrule
\end{tabular}

    }
    \caption{Average Accuracy (\%; higher is better) on four benchmark datasets with different contamination ratios $P$. Numbers in parentheses indicate the performance degradation due to contamination compared to the clean setting. The average and deviation over 10 runs are reported.}
    \label{tab:mix}
    \vspace{-8pt}
\end{table*}

\begin{table*}
    \centering
    \resizebox{.7\textwidth}{!}{
    \begin{tabular}{lcccccc}
\toprule
\multicolumn{1}{c}{Dataset} & CIFAR10 & \multicolumn{5}{c}{C10/SDXL}\\
\cmidrule(lr){1-1}
\cmidrule(lr){2-2}
\cmidrule(lr){3-7}

\multicolumn{1}{c}{Ratio $P$} & 0\% & 50\% & 70\% & 80\% & 90\% & 95\% \\
\midrule

ER 	&	80.05{\scriptsize ±3.24}&	79.56{\scriptsize ±2.51}&	74.76{\scriptsize ±3.04}&	74.15{\scriptsize ±3.80}&	72.69{\scriptsize ±3.02}&	69.60{\scriptsize ±4.38}\\
DER++ &	78.35{\scriptsize ±1.88}&	74.51{\scriptsize ±3.63}&	71.33{\scriptsize ±4.16}&	67.68{\scriptsize ±4.23}&	62.96{\scriptsize ±2.89}&	60.45{\scriptsize ±3.05}\\
ERACE &	60.64{\scriptsize ±3.24}&	57.16{\scriptsize ±3.18}&	49.61{\scriptsize ±5.13}&	46.44{\scriptsize ±5.47}&	43.70{\scriptsize ±3.48}&	40.94{\scriptsize ±4.11}\\
OCM 	&	79.84{\scriptsize ±3.01}&	79.25{\scriptsize ±2.28}&	76.43{\scriptsize ±3.26}&	76.03{\scriptsize ±3.39}&	74.41{\scriptsize ±2.51}&	71.11{\scriptsize ±1.94}\\
GSA 	&	77.28{\scriptsize ±3.11}&	74.72{\scriptsize ±3.39}&	69.10{\scriptsize ±2.65}&	66.51{\scriptsize ±3.92}&	60.17{\scriptsize ±4.53}&	56.57{\scriptsize ±7.26}\\
OnPro &	84.87{\scriptsize ±2.55}&	83.12{\scriptsize ±3.30}&	82.98{\scriptsize ±1.75}&	79.81{\scriptsize ±2.82}&	77.99{\scriptsize ±2.88}&	73.39{\scriptsize ±3.26}\\
ESRM 	&	91.36{\scriptsize ±1.25}&	88.19{\scriptsize ±1.05}&	85.57{\scriptsize ±1.59}&	83.32{\scriptsize ±1.70}&	79.48{\scriptsize ±2.09}&	76.09{\scriptsize ±3.10}\\

\midrule
\multicolumn{1}{c}{Dataset} & CIFAR100 & \multicolumn{5}{c}{C100/SDXL}\\
\cmidrule(lr){1-1}
\cmidrule(lr){2-2}
\cmidrule(lr){3-7}

\multicolumn{1}{c}{Ratio $P$} & 0\% & 50\% & 70\% & 80\% & 90\% & 95\% \\
\midrule

ER 	&	50.47{\scriptsize ±1.31}&	48.60{\scriptsize ±1.77}&	46.08{\scriptsize ±1.69}&	43.19{\scriptsize ±1.28}&	39.66{\scriptsize ±1.47}&	37.21{\scriptsize ±1.49}\\
DER++ &	55.91{\scriptsize ±3.80}&	50.80{\scriptsize ±2.52}&	46.42{\scriptsize ±2.86}&	43.17{\scriptsize ±2.72}&	36.79{\scriptsize ±3.14}&	31.44{\scriptsize ±2.20}\\
ERACE &	41.29{\scriptsize ±1.72}&	35.83{\scriptsize ±0.47}&	30.06{\scriptsize ±1.22}&	26.56{\scriptsize ±1.24}&	21.70{\scriptsize ±1.64}&	18.10{\scriptsize ±0.68}\\
OCM 	&	43.56{\scriptsize ±1.56}&	43.30{\scriptsize ±1.86}&	40.59{\scriptsize ±1.52}&	38.62{\scriptsize ±1.94}&	35.73{\scriptsize ±1.57}&	33.93{\scriptsize ±1.81}\\
GSA 	&	50.76{\scriptsize ±1.64}&	48.30{\scriptsize ±2.28}&	43.70{\scriptsize ±2.17}&	40.49{\scriptsize ±1.75}&	34.29{\scriptsize ±1.48}&	30.92{\scriptsize ±1.56}\\
OnPro &	41.98{\scriptsize ±1.44}&	41.58{\scriptsize ±1.85}&	39.36{\scriptsize ±1.36}&	36.99{\scriptsize ±1.87}&	33.95{\scriptsize ±1.61}&	30.97{\scriptsize ±1.81}\\
ESRM 	&	70.85{\scriptsize ±0.95}&	67.07{\scriptsize ±0.77}&	62.83{\scriptsize ±0.64}&	59.45{\scriptsize ±1.00}&	54.33{\scriptsize ±0.52}&	49.89{\scriptsize ±0.78}\\

\midrule
\multicolumn{1}{c}{Dataset} & Tiny & \multicolumn{5}{c}{Tiny/SDXL}\\
\cmidrule(lr){1-1}
\cmidrule(lr){2-2}
\cmidrule(lr){3-7}

\multicolumn{1}{c}{Ratio $P$} & 0\% & 50\% & 70\% & 80\% & 90\% & 95\% \\
\midrule

ER 	&	64.44{\scriptsize ±1.27}&	56.76{\scriptsize ±1.23}&	51.90{\scriptsize ±1.95}&	47.26{\scriptsize ±0.94}&	36.63{\scriptsize ±1.39}&	24.89{\scriptsize ±1.23}\\
DER++ &	70.28{\scriptsize ±2.42}&	64.84{\scriptsize ±1.48}&	61.25{\scriptsize ±2.79}&	58.82{\scriptsize ±1.92}&	54.52{\scriptsize ±1.88}&	49.59{\scriptsize ±1.56}\\
ERACE &	4.60{\scriptsize ±0.88}&	3.50{\scriptsize ±0.45}&	3.00{\scriptsize ±0.50}&	2.80{\scriptsize ±0.68}&	2.24{\scriptsize ±0.35}&	1.96{\scriptsize ±0.34}\\
OCM 	&	14.91{\scriptsize ±2.23}&	9.34{\scriptsize ±1.27}&	7.49{\scriptsize ±1.12}&	6.65{\scriptsize ±0.94}&	5.43{\scriptsize ±1.04}&	4.56{\scriptsize ±0.89}\\
GSA 	&	14.95{\scriptsize ±0.52}&	8.67{\scriptsize ±1.38}&	6.68{\scriptsize ±0.62}&	5.06{\scriptsize ±0.61}&	3.19{\scriptsize ±0.82}&	2.37{\scriptsize ±0.36}\\
OnPro &	15.82{\scriptsize ±1.04}&	11.18{\scriptsize ±1.40}&	9.45{\scriptsize ±1.62}&	8.68{\scriptsize ±0.72}&	8.30{\scriptsize ±1.30}&	6.75{\scriptsize ±1.36}\\
ESRM 	&	81.47{\scriptsize ±0.58}&	73.76{\scriptsize ±0.44}&	66.65{\scriptsize ±0.64}&	60.88{\scriptsize ±0.96}&	50.58{\scriptsize ±1.15}&	38.43{\scriptsize ±0.91}\\

\midrule
\multicolumn{1}{c}{Dataset} & CIFAR10 & \multicolumn{5}{c}{C10/Mix}\\
\cmidrule(lr){1-1}
\cmidrule(lr){2-2}
\cmidrule(lr){3-7}

\multicolumn{1}{c}{Ratio $P$} & 0\% & 50\% & 70\% & 80\% & 90\% & 95\% \\
\midrule

ER 	&	80.05{\scriptsize ±3.24}&	77.03{\scriptsize ±3.53}&	75.06{\scriptsize ±3.60}&	74.57{\scriptsize ±3.31}&	72.41{\scriptsize ±3.11}&	71.68{\scriptsize ±3.89}\\
DER++ &	78.35{\scriptsize ±1.88}&	75.53{\scriptsize ±2.35}&	74.16{\scriptsize ±3.44}&	70.95{\scriptsize ±2.93}&	69.57{\scriptsize ±2.82}&	67.31{\scriptsize ±2.27}\\
ERACE &	60.64{\scriptsize ±3.24}&	57.57{\scriptsize ±4.67}&	52.87{\scriptsize ±3.40}&	49.89{\scriptsize ±4.71}&	49.55{\scriptsize ±4.95}&	46.53{\scriptsize ±4.00}\\
OCM 	&	79.84{\scriptsize ±3.01}&	78.75{\scriptsize ±3.58}&	76.91{\scriptsize ±2.67}&	75.78{\scriptsize ±2.84}&	73.60{\scriptsize ±2.72}&	72.36{\scriptsize ±2.44}\\
GSA 	&	77.28{\scriptsize ±3.11}&	72.38{\scriptsize ±2.99}&	72.11{\scriptsize ±3.74}&	69.66{\scriptsize ±2.58}&	64.43{\scriptsize ±4.81}&	60.80{\scriptsize ±4.32}\\
OnPro &	84.87{\scriptsize ±2.55}&	82.96{\scriptsize ±2.97}&	80.94{\scriptsize ±2.79}&	79.00{\scriptsize ±2.31}&	76.94{\scriptsize ±2.75}&	76.17{\scriptsize ±3.31}\\
ESRM 	&	91.36{\scriptsize ±1.25}&	88.81{\scriptsize ±1.49}&	85.98{\scriptsize ±1.44}&	84.60{\scriptsize ±1.96}&	82.62{\scriptsize ±0.82}&	81.11{\scriptsize ±1.58}\\

\midrule
\multicolumn{1}{c}{Dataset} & CIFAR100 & \multicolumn{5}{c}{C100/Mix}\\
\cmidrule(lr){1-1}
\cmidrule(lr){2-2}
\cmidrule(lr){3-7}

\multicolumn{1}{c}{Ratio $P$} & 0\% & 50\% & 70\% & 80\% & 90\% & 95\% \\
\midrule

ER 	&	50.47{\scriptsize ±1.31}&	48.64{\scriptsize ±1.91}&	45.45{\scriptsize ±1.37}&	43.30{\scriptsize ±1.20}&	41.36{\scriptsize ±1.29}&	39.28{\scriptsize ±1.74}\\
DER++ &	55.91{\scriptsize ±3.80}&	54.30{\scriptsize ±3.50}&	50.02{\scriptsize ±1.74}&	47.44{\scriptsize ±2.68}&	43.39{\scriptsize ±1.61}&	41.68{\scriptsize ±2.21}\\
ERACE &	41.29{\scriptsize ±1.72}&	37.41{\scriptsize ±1.00}&	33.96{\scriptsize ±1.31}&	31.89{\scriptsize ±1.16}&	28.65{\scriptsize ±1.07}&	26.62{\scriptsize ±0.89}\\
OCM 	&	43.56{\scriptsize ±1.56}&	42.41{\scriptsize ±2.05}&	39.26{\scriptsize ±1.65}&	38.16{\scriptsize ±1.90}&	37.45{\scriptsize ±1.48}&	36.16{\scriptsize ±1.47}\\
GSA 	&	50.76{\scriptsize ±1.64}&	48.07{\scriptsize ±1.31}&	45.37{\scriptsize ±1.44}&	42.83{\scriptsize ±1.59}&	39.65{\scriptsize ±1.21}&	37.85{\scriptsize ±1.41}\\
OnPro &	41.98{\scriptsize ±1.44}&	41.62{\scriptsize ±1.46}&	39.24{\scriptsize ±1.53}&	36.95{\scriptsize ±1.60}&	36.02{\scriptsize ±1.01}&	34.94{\scriptsize ±1.11}\\
ESRM 	&	70.85{\scriptsize ±0.95}&	67.53{\scriptsize ±1.08}&	64.04{\scriptsize ±0.91}&	61.34{\scriptsize ±1.01}&	57.57{\scriptsize ±0.73}&	54.76{\scriptsize ±0.75}\\

\midrule
\multicolumn{1}{c}{Dataset} & Tiny & \multicolumn{5}{c}{Tiny/Mix}\\
\cmidrule(lr){1-1}
\cmidrule(lr){2-2}
\cmidrule(lr){3-7}

\multicolumn{1}{c}{Ratio $P$} & 0\% & 50\% & 70\% & 80\% & 90\% & 95\% \\
\midrule

ER 	&	64.44{\scriptsize ±1.27}&	55.16{\scriptsize ±1.06}&	50.39{\scriptsize ±2.59}&	44.24{\scriptsize ±2.08}&	32.22{\scriptsize ±1.74}&	22.06{\scriptsize ±1.74}\\
DER++ &	70.28{\scriptsize ±2.42}&	64.64{\scriptsize ±1.19}&	62.77{\scriptsize ±2.16}&	60.53{\scriptsize ±2.12}&	53.28{\scriptsize ±2.80}&	46.38{\scriptsize ±2.73}\\
ERACE &	4.60{\scriptsize ±0.88}&	3.04{\scriptsize ±0.75}&	2.63{\scriptsize ±0.55}&	2.29{\scriptsize ±0.48}&	1.87{\scriptsize ±0.37}&	1.86{\scriptsize ±0.25}\\
OCM 	&	14.91{\scriptsize ±2.23}&	8.58{\scriptsize ±1.20}&	6.18{\scriptsize ±1.17}&	4.72{\scriptsize ±1.03}&	4.01{\scriptsize ±0.75}&	3.42{\scriptsize ±0.63}\\
GSA 	&	14.95{\scriptsize ±0.52}&	8.66{\scriptsize ±1.18}&	6.53{\scriptsize ±1.00}&	4.90{\scriptsize ±0.66}&	2.84{\scriptsize ±0.58}&	2.06{\scriptsize ±0.49}\\
OnPro &	15.82{\scriptsize ±1.04}&	9.83{\scriptsize ±1.06}&	7.92{\scriptsize ±1.22}&	6.89{\scriptsize ±1.13}&	5.62{\scriptsize ±0.85}&	4.32{\scriptsize ±1.28}\\
ESRM 	&	81.47{\scriptsize ±0.58}&	74.50{\scriptsize ±0.59}&	66.60{\scriptsize ±1.00}&	60.56{\scriptsize ±0.75}&	48.22{\scriptsize ±1.16}&	34.33{\scriptsize ±1.12}\\

\bottomrule
\end{tabular}
    }
    \caption{Learning Accuracy (\%; higher is better) on contaminated datasets with various contamination ratio $P$.}
    \label{tab:laext}
    \vspace{-8pt}
\end{table*}

\begin{table*}
    \centering
    \resizebox{.7\textwidth}{!}{
    \begin{tabular}{lcccccc}
\toprule
\multicolumn{1}{c}{Dataset} & CIFAR10 & \multicolumn{5}{c}{C10/SDXL}\\
\cmidrule(lr){1-1}
\cmidrule(lr){2-2}
\cmidrule(lr){3-7}

\multicolumn{1}{c}{Ratio $P$} & 0\% & 50\% & 70\% & 80\% & 90\% & 95\% \\
\midrule

ER 	&	20.30{\scriptsize ±4.50}&	23.47{\scriptsize ±3.80}&	23.56{\scriptsize ±5.66}&	25.71{\scriptsize ±5.14}&	29.74{\scriptsize ±5.20}&	30.08{\scriptsize ±6.21}\\
DER++ &	17.36{\scriptsize ±2.62}&	18.64{\scriptsize ±3.57}&	19.65{\scriptsize ±2.64}&	21.46{\scriptsize ±3.63}&	25.74{\scriptsize ±4.63}&	28.72{\scriptsize ±4.44}\\
ERACE &	9.59{\scriptsize ±2.12}&	10.69{\scriptsize ±1.30}&	13.97{\scriptsize ±3.22}&	15.83{\scriptsize ±4.33}&	15.70{\scriptsize ±3.72}&	20.51{\scriptsize ±2.05}\\
OCM 	&	10.43{\scriptsize ±2.51}&	12.44{\scriptsize ±1.72}&	13.43{\scriptsize ±3.90}&	16.39{\scriptsize ±3.01}&	18.21{\scriptsize ±2.95}&	18.68{\scriptsize ±3.39}\\
GSA 	&	16.25{\scriptsize ±2.62}&	17.41{\scriptsize ±2.92}&	16.50{\scriptsize ±3.44}&	17.10{\scriptsize ±4.02}&	21.34{\scriptsize ±5.05}&	21.12{\scriptsize ±7.01}\\
OnPro &	12.07{\scriptsize ±3.19}&	13.05{\scriptsize ±2.64}&	15.65{\scriptsize ±2.96}&	16.88{\scriptsize ±2.16}&	19.12{\scriptsize ±4.23}&	21.94{\scriptsize ±5.57}\\
ESRM 	&	26.12{\scriptsize ±2.05}&	23.03{\scriptsize ±2.16}&	21.27{\scriptsize ±1.96}&	20.24{\scriptsize ±2.42}&	20.52{\scriptsize ±2.70}&	23.99{\scriptsize ±3.23}\\

\midrule
\multicolumn{1}{c}{Dataset} & CIFAR100 & \multicolumn{5}{c}{C100/SDXL}\\
\cmidrule(lr){1-1}
\cmidrule(lr){2-2}
\cmidrule(lr){3-7}

\multicolumn{1}{c}{Ratio $P$} & 0\% & 50\% & 70\% & 80\% & 90\% & 95\% \\
\midrule

ER 	&	23.86{\scriptsize ±2.73}&	26.06{\scriptsize ±4.53}&	28.51{\scriptsize ±3.56}&	27.34{\scriptsize ±3.41}&	31.48{\scriptsize ±2.95}&	30.91{\scriptsize ±3.12}\\
DER++ &	33.11{\scriptsize ±4.58}&	35.73{\scriptsize ±5.20}&	38.56{\scriptsize ±4.77}&	39.97{\scriptsize ±5.72}&	46.20{\scriptsize ±4.12}&	47.89{\scriptsize ±5.86}\\
ERACE &	9.81{\scriptsize ±1.72}&	12.26{\scriptsize ±1.91}&	13.86{\scriptsize ±2.75}&	14.15{\scriptsize ±3.64}&	18.54{\scriptsize ±5.33}&	19.61{\scriptsize ±5.17}\\
OCM 	&	8.35{\scriptsize ±2.07}&	10.46{\scriptsize ±2.62}&	10.93{\scriptsize ±3.57}&	12.61{\scriptsize ±1.80}&	15.41{\scriptsize ±2.48}&	17.60{\scriptsize ±3.00}\\
GSA 	&	19.62{\scriptsize ±2.73}&	21.56{\scriptsize ±3.38}&	22.69{\scriptsize ±2.95}&	23.05{\scriptsize ±4.13}&	23.45{\scriptsize ±3.00}&	25.38{\scriptsize ±4.53}\\
OnPro &	11.91{\scriptsize ±1.87}&	13.28{\scriptsize ±2.40}&	15.99{\scriptsize ±1.99}&	16.58{\scriptsize ±2.13}&	17.01{\scriptsize ±2.50}&	18.78{\scriptsize ±3.58}\\
ESRM 	&	32.67{\scriptsize ±1.46}&	30.75{\scriptsize ±1.48}&	26.93{\scriptsize ±1.33}&	25.24{\scriptsize ±1.64}&	24.53{\scriptsize ±1.04}&	24.85{\scriptsize ±1.77}\\

\midrule
\multicolumn{1}{c}{Dataset} & Tiny & \multicolumn{5}{c}{Tiny/SDXL}\\
\cmidrule(lr){1-1}
\cmidrule(lr){2-2}
\cmidrule(lr){3-7}

\multicolumn{1}{c}{Ratio $P$} & 0\% & 50\% & 70\% & 80\% & 90\% & 95\% \\
\midrule

ER   &	61.84{\scriptsize ±2.65}&	69.08{\scriptsize ±3.09}&	74.91{\scriptsize ±3.41}&	76.94{\scriptsize ±2.89}&	83.45{\scriptsize ±2.34}&	86.34{\scriptsize ±2.27}\\
DER++ &	72.51{\scriptsize ±5.53}&	80.83{\scriptsize ±3.51}&	84.35{\scriptsize ±2.27}&	87.53{\scriptsize ±2.58}&	91.88{\scriptsize ±2.09}&	94.46{\scriptsize ±0.72}\\
ERACE &	36.40{\scriptsize ±2.74}&	44.92{\scriptsize ±2.49}&	54.44{\scriptsize ±3.07}&	62.57{\scriptsize ±5.27}&	76.19{\scriptsize ±3.48}&	78.71{\scriptsize ±4.21}\\
OCM  &	32.25{\scriptsize ±1.44}&	37.15{\scriptsize ±1.73}&	43.82{\scriptsize ±2.82}&	49.69{\scriptsize ±2.54}&	58.53{\scriptsize ±2.89}&	73.77{\scriptsize ±4.01}\\
GSA  &	44.78{\scriptsize ±2.76}&	55.72{\scriptsize ±5.38}&	58.39{\scriptsize ±3.46}&	64.32{\scriptsize ±3.04}&	71.39{\scriptsize ±3.42}&	76.41{\scriptsize ±4.15}\\
OnPro &	42.81{\scriptsize ±4.63}&	52.63{\scriptsize ±4.17}&	56.83{\scriptsize ±3.74}&	65.18{\scriptsize ±3.77}&	75.11{\scriptsize ±5.65}&	80.98{\scriptsize ±1.55}\\
ESRM  &	61.53{\scriptsize ±1.36}&	61.35{\scriptsize ±0.66}&	59.79{\scriptsize ±1.82}&	59.55{\scriptsize ±2.17}&	65.44{\scriptsize ±2.72}&	71.04{\scriptsize ±2.10}\\

\midrule
\multicolumn{1}{c}{Dataset} & CIFAR10 & \multicolumn{5}{c}{C10/Mix}\\
\cmidrule(lr){1-1}
\cmidrule(lr){2-2}
\cmidrule(lr){3-7}

\multicolumn{1}{c}{Ratio $P$} & 0\% & 50\% & 70\% & 80\% & 90\% & 95\% \\
\midrule

ER   &	20.30{\scriptsize ±4.50}&	20.20{\scriptsize ±3.16}&	22.77{\scriptsize ±4.62}&	22.05{\scriptsize ±5.46}&	25.60{\scriptsize ±3.91}&	26.89{\scriptsize ±4.03}\\
DER++ &	17.36{\scriptsize ±2.62}&	17.76{\scriptsize ±2.03}&	19.10{\scriptsize ±3.82}&	18.49{\scriptsize ±3.86}&	21.71{\scriptsize ±7.22}&	20.94{\scriptsize ±4.21}\\
ERACE &	9.59{\scriptsize ±2.12}&	10.67{\scriptsize ±2.66}&	11.61{\scriptsize ±2.12}&	13.58{\scriptsize ±4.21}&	14.09{\scriptsize ±4.70}&	12.06{\scriptsize ±2.61}\\
OCM  &	10.43{\scriptsize ±2.51}&	11.04{\scriptsize ±2.82}&	12.89{\scriptsize ±3.58}&	14.24{\scriptsize ±3.79}&	15.01{\scriptsize ±2.95}&	15.15{\scriptsize ±2.33}\\
GSA  &	16.25{\scriptsize ±2.62}&	15.33{\scriptsize ±4.76}&	16.34{\scriptsize ±3.71}&	16.28{\scriptsize ±3.63}&	15.94{\scriptsize ±2.53}&	16.46{\scriptsize ±2.97}\\
OnPro &	12.07{\scriptsize ±3.19}&	12.72{\scriptsize ±2.70}&	12.93{\scriptsize ±3.08}&	13.39{\scriptsize ±3.11}&	14.36{\scriptsize ±2.92}&	14.49{\scriptsize ±3.65}\\
ESRM  &	26.12{\scriptsize ±2.05}&	23.75{\scriptsize ±2.39}&	21.81{\scriptsize ±2.53}&	21.83{\scriptsize ±2.24}&	22.77{\scriptsize ±1.15}&	23.48{\scriptsize ±2.78}\\

\midrule
\multicolumn{1}{c}{Dataset} & CIFAR100 & \multicolumn{5}{c}{C100/Mix}\\
\cmidrule(lr){1-1}
\cmidrule(lr){2-2}
\cmidrule(lr){3-7}

\multicolumn{1}{c}{Ratio $P$} & 0\% & 50\% & 70\% & 80\% & 90\% & 95\% \\
\midrule

ER   &	23.86{\scriptsize ±2.73}&	25.47{\scriptsize ±3.32}&	25.92{\scriptsize ±3.13}&	25.18{\scriptsize ±2.30}&	26.70{\scriptsize ±2.79}&	27.95{\scriptsize ±3.58}\\
DER++ &	33.11{\scriptsize ±4.58}&	34.53{\scriptsize ±3.63}&	36.41{\scriptsize ±3.53}&	36.96{\scriptsize ±4.55}&	37.83{\scriptsize ±2.08}&	39.97{\scriptsize ±2.38}\\
ERACE &	9.81{\scriptsize ±1.72}&	11.32{\scriptsize ±1.66}&	14.62{\scriptsize ±3.75}&	15.33{\scriptsize ±3.12}&	16.16{\scriptsize ±3.43}&	18.34{\scriptsize ±3.36}\\
OCM  &	8.35{\scriptsize ±2.07}&	9.87{\scriptsize ±2.31}&	9.98{\scriptsize ±2.21}&	11.12{\scriptsize ±3.01}&	13.87{\scriptsize ±3.72}&	14.54{\scriptsize ±3.62}\\
GSA  &	19.62{\scriptsize ±2.73}&	20.00{\scriptsize ±1.83}&	20.71{\scriptsize ±1.27}&	22.26{\scriptsize ±2.97}&	22.06{\scriptsize ±3.03}&	22.01{\scriptsize ±4.44}\\
OnPro &	11.91{\scriptsize ±1.87}&	14.28{\scriptsize ±2.69}&	13.79{\scriptsize ±1.32}&	14.50{\scriptsize ±2.16}&	15.71{\scriptsize ±3.13}&	16.36{\scriptsize ±4.01}\\
ESRM  &	32.67{\scriptsize ±1.46}&	32.72{\scriptsize ±1.77}&	31.34{\scriptsize ±1.26}&	30.19{\scriptsize ±2.04}&	29.94{\scriptsize ±1.34}&	30.19{\scriptsize ±1.70}\\

\midrule
\multicolumn{1}{c}{Dataset} & Tiny & \multicolumn{5}{c}{Tiny/Mix}\\
\cmidrule(lr){1-1}
\cmidrule(lr){2-2}
\cmidrule(lr){3-7}

\multicolumn{1}{c}{Ratio $P$} & 0\% & 50\% & 70\% & 80\% & 90\% & 95\% \\
\midrule

ER   &	61.84{\scriptsize ±2.65}&	69.86{\scriptsize ±3.33}&	74.38{\scriptsize ±2.97}&	77.56{\scriptsize ±3.09}&	81.40{\scriptsize ±3.07}&	87.49{\scriptsize ±3.14}\\
DER++ &	72.51{\scriptsize ±5.53}&	78.64{\scriptsize ±4.06}&	83.17{\scriptsize ±1.41}&	87.87{\scriptsize ±3.32}&	93.01{\scriptsize ±1.55}&	96.49{\scriptsize ±1.26}\\
ERACE &	36.40{\scriptsize ±2.74}&	43.29{\scriptsize ±2.72}&	48.53{\scriptsize ±3.15}&	55.66{\scriptsize ±3.50}&	66.81{\scriptsize ±3.03}&	74.77{\scriptsize ±3.71}\\
OCM  &	32.25{\scriptsize ±1.44}&	38.31{\scriptsize ±2.25}&	45.04{\scriptsize ±3.00}&	49.95{\scriptsize ±2.56}&	61.10{\scriptsize ±1.94}&	74.39{\scriptsize ±4.50}\\
GSA  &	44.78{\scriptsize ±2.76}&	51.89{\scriptsize ±4.09}&	57.09{\scriptsize ±5.29}&	63.64{\scriptsize ±4.54}&	71.41{\scriptsize ±3.71}&	81.18{\scriptsize ±4.13}\\
OnPro &	42.81{\scriptsize ±4.63}&	52.02{\scriptsize ±3.70}&	58.82{\scriptsize ±3.49}&	63.49{\scriptsize ±4.40}&	76.68{\scriptsize ±3.02}&	82.61{\scriptsize ±6.06}\\
ESRM  &	61.53{\scriptsize ±1.36}&	64.21{\scriptsize ±1.54}&	62.78{\scriptsize ±1.66}&	62.82{\scriptsize ±2.15}&	67.37{\scriptsize ±2.25}&	71.86{\scriptsize ±2.36}\\

\bottomrule
\end{tabular}
    }
    \caption{Relative Forgetting (\%; lower is better) on contaminated datasets with various contamination ratio $P$.}
    \label{tab:rfext}
    \vspace{-8pt}
\end{table*}

\begin{table*}
    \centering
    \resizebox{\textwidth}{!}{
    \begin{tabular}{lcccccc}
\toprule
\multicolumn{1}{c}{Dataset} & C100(M=1k) & \multicolumn{5}{c}{C100/SDXL (M=1k)}\\
\cmidrule(lr){1-1}
\cmidrule(lr){2-2}
\cmidrule(lr){3-7}

\multicolumn{1}{c}{Ratio $P$} & 0\% & 50\% & 70\% & 80\% & 90\% & 95\% \\
\midrule

ER   &	24.92{\scriptsize ±1.33} &	23.25{\scriptsize ±0.98}(-1.67) &	22.18{\scriptsize ±1.58}(-2.74) &	20.63{\scriptsize ±1.27}(-4.29) &	18.96{\scriptsize ±1.54}(-5.96) &	17.79{\scriptsize ±0.97}(-7.13) \\	
DER++ &	25.86{\scriptsize ±2.43} &	21.92{\scriptsize ±1.36}(-3.94) &	19.47{\scriptsize ±1.14}(-6.39) &	16.69{\scriptsize ±1.65}(-9.17) &	13.60{\scriptsize ±1.06}(-12.26) &	11.15{\scriptsize ±1.31}(-14.71) \\	
ERACE &	\textbf{28.22{\scriptsize ±1.09}} &	23.96{\scriptsize ±0.92}(-4.26) &	19.38{\scriptsize ±0.78}(-8.84) &	17.71{\scriptsize ±0.90}(-10.51) &	13.85{\scriptsize ±0.71}(-14.37) &	11.94{\scriptsize ±0.43}(-16.28) \\	
OCM  &	28.02{\scriptsize ±0.74} &	26.54{\scriptsize ±1.02}(-1.48) &	24.94{\scriptsize ±1.07}(-3.08) &	24.17{\scriptsize ±1.03}(-3.85) &	22.12{\scriptsize ±0.64}(-5.90) &	20.65{\scriptsize ±1.08}(-7.37) \\
GSA  &	28.15{\scriptsize ±1.59} &	26.20{\scriptsize ±1.47}(-1.95) &	23.52{\scriptsize ±1.14}(-4.63) &	22.90{\scriptsize ±0.91}(-5.25) &	19.20{\scriptsize ±1.27}(-8.95) &	16.89{\scriptsize ±1.37}(-11.26) \\	
OnPro &	26.92{\scriptsize ±1.31} &	26.31{\scriptsize ±1.30}(-0.61) &	24.85{\scriptsize ±0.94}(-2.07) &	23.10{\scriptsize ±1.34}(-3.82) &	20.98{\scriptsize ±1.02}(-5.94) &	19.77{\scriptsize ±1.49}(-7.15) \\
ESRM  &	27.14{\scriptsize ±0.79} &	\textbf{26.57{\scriptsize ±1.10}(-0.57)} &	\textbf{26.17{\scriptsize ±0.91}(-0.97)} &	\textbf{25.05{\scriptsize ±0.50}(-2.09)} &	\textbf{24.43{\scriptsize ±0.91}(-2.71)} &	\textbf{23.35{\scriptsize ±1.17}(-3.79)} \\

\midrule
\multicolumn{1}{c}{Dataset} & C100(M=2k) & \multicolumn{5}{c}{C100/SDXL (M=2K)}\\
\cmidrule(lr){1-1}
\cmidrule(lr){2-2}
\cmidrule(lr){3-7}

\multicolumn{1}{c}{Ratio $P$} & 0\% & 50\% & 70\% & 80\% & 90\% & 95\% \\
\midrule

ER   &	32.07{\scriptsize ±1.51} &	30.24{\scriptsize ±1.15}(-1.83) &	27.38{\scriptsize ±1.41}(-4.69) &	25.95{\scriptsize ±1.22}(-6.12) &	23.68{\scriptsize ±1.14}(-8.39) &	21.79{\scriptsize ±0.65}(-10.28) \\
DER++ &	33.37{\scriptsize ±2.11} &	27.95{\scriptsize ±2.12}(-5.42) &	24.60{\scriptsize ±1.53}(-8.77) &	21.07{\scriptsize ±1.89}(-12.30) &	17.39{\scriptsize ±1.06}(-15.98) &	14.15{\scriptsize ±1.52}(-19.22) \\
ERACE &	34.30{\scriptsize ±1.49} &	28.69{\scriptsize ±1.71}(-5.61) &	24.01{\scriptsize ±1.14}(-10.29) &	21.23{\scriptsize ±0.84}(-13.07) &	16.47{\scriptsize ±0.95}(-17.83) &	13.25{\scriptsize ±1.68}(-21.05) \\
OCM  &	35.69{\scriptsize ±1.36} &	32.39{\scriptsize ±1.09}(-3.30) &	31.15{\scriptsize ±0.84}(-4.54) &	28.38{\scriptsize ±1.28}(-7.31) &	26.76{\scriptsize ±0.79}(-8.93) &	24.75{\scriptsize ±0.67}(-10.94) \\
GSA  &	35.31{\scriptsize ±1.47} &	32.72{\scriptsize ±1.33}(-2.59) &	28.97{\scriptsize ±1.23}(-6.34) &	26.87{\scriptsize ±1.03}(-8.44) &	23.42{\scriptsize ±1.17}(-11.89) &	19.80{\scriptsize ±1.32}(-15.51) \\
OnPro &	33.52{\scriptsize ±0.80} &	31.33{\scriptsize ±0.75}(-2.19) &	30.02{\scriptsize ±1.01}(-3.50) &	27.90{\scriptsize ±0.85}(-5.62) &	24.38{\scriptsize ±0.69}(-9.14) &	22.58{\scriptsize ±1.18}(-10.94) \\
ESRM  &	\textbf{36.25{\scriptsize ±0.79}} &	\textbf{34.55{\scriptsize ±1.33}(-1.70)} &	\textbf{34.13{\scriptsize ±1.03}(-2.12)} &	\textbf{33.77{\scriptsize ±1.08}(-2.48)} &	\textbf{31.70{\scriptsize ±0.86}(-4.55)} &	\textbf{29.21{\scriptsize ±0.91}(-7.04)} \\

\bottomrule
\end{tabular}

    }
    \caption{Final Average Accuracy (\%; higher is better) on C100/SDXL dataset with different memory size $M$ and contamination ratio $P$. Numbers in parentheses indicate the performance degradation due to contamination compared to the clean setting. The average and deviation over 10 runs are reported.}
    \label{tab:memc100}
    \vspace{-8pt}
\end{table*}

\begin{figure}
    \centering
    \includegraphics[width=\linewidth]{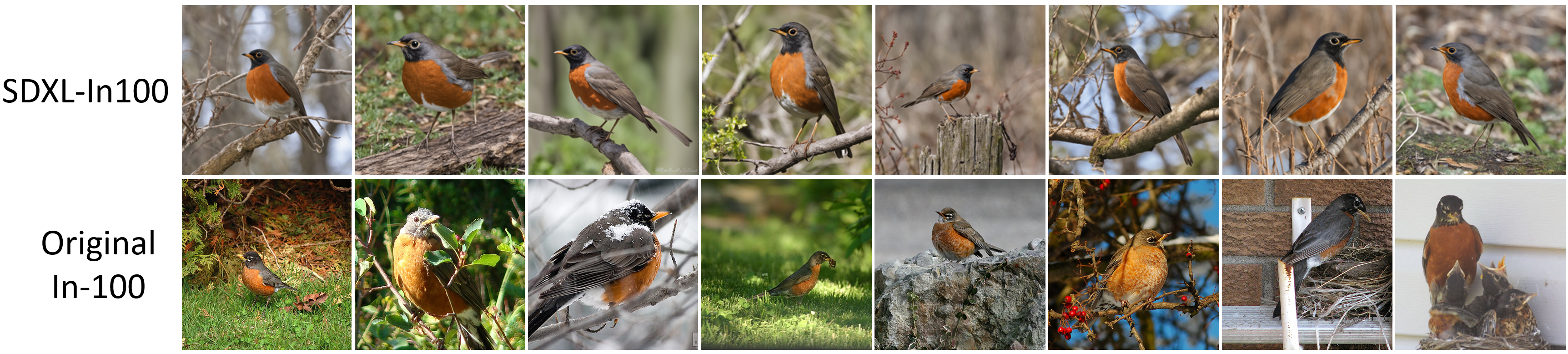}
    \caption{Random sampled images from class ``n01558993'' (Robin) in SDXL-In100 and original ImageNet-100 dataset. For clarity, we have cropped some backgrounds and resized the vanilla ImageNet-100 samples.}
    \label{fig:samples}
\end{figure}

\section{Implementation Details}
\label{apdx:implementation}

\subsection{Dataset.}
\label{apdx:dataset}
As discussed in Sec.~\ref{sec:preliminary}, we used four benchmark datasets in evaluation, including CIFAR-10/100, TinyImageNet, and ImageNet-100. In the experiments, all of the datasets are split into tasks containing non-overlapping classes. The details about the task split are as follows:

\textbf{CIFAR-10}~\cite{CIFAR} has ten classes with 50,000 training images and 10,000 test images. The image is $32 \times 32$ in size. The dataset is split into five disjoint tasks with two classes per task.

\textbf{CIFAR-100}~\cite{CIFAR} has 100 classes with 50,000 training samples and 10,000 test samples. Image size is $32 \times 32$. It is split into 10 disjoint tasks with 10 classes per task.

\textbf{TinyImageNet}~\cite{Tiny} has 200 classes, 100,000 training samples, and 10,000 test samples. Image size is  $64 \times 64$. The dataset is split into 100 non-overlapping tasks with two classes per task.

\textbf{ImageNet-100}~\cite{In100} is a subset of ImageNet-1k~\cite{Imagenet} dataset. It consists of 100 classes. We follow~\cite{PyIN100} for the class selection. We do not perform image resizing in generating ImageNet-100 from the original ImageNet-1k dataset. The dataset is split into 10 disjoint classes with 10 classes per task.

\subsection{Details about synthetic dataset generation.}
\label{apdx:generation}
\textbf{Image size of SDXL-IN100.} In the synthetic dataset generation, we manually adjust the size of generated images to match that of the original dataset. For the ImageNet-100 dataset, since the image size is not fixed, we resize the generated images to $224 \times 224$, to align with the training protocols.

\textbf{Class-wise distribution of sources for Mix-C10/C100/Tiny dataset.} In the main manuscript, we mentioned that the dataset of setting \textbf{b)} in Sec.~\ref{subsec:syn_gen} is generated from five synthetic models: Stable Diffusion v1.4, Stable Diffusion v2.1, Stable Diffusion XL, VQDM, and GLIDE. Each method contributes 20\% of the dataset in setting \textbf{b)}. In our implementation, we ensure that this distribution is consistent across all classes in the dataset, so that each class has an equal number of images from each generation model.

\begin{table}[t]
    \centering
    \resizebox{.7\linewidth}{!}{
    
\begin{tabular}{cccc}
\toprule
Generative Model&Diffusion Steps&Upsample(Refiner) Steps&Guidance Scale\\
\midrule
SD1.4	&50	    &N/A	&7.5\\
SD2.1	&50	    &N/A	&7.5\\
SDXL	&40	    &40	&5.0\\
VQDM	&100	&N/A	&7.5\\
GLIDE	&100	&27	&3.0\\
\bottomrule
\end{tabular}
    }
    \vspace{5pt}
    \caption{Hyperparameters used in image generation.}
    \label{tab:DMParams}
\end{table}

\textbf{Hyperparameters used in image generation.} For Stable Diffusion and VQDM, we use source code and model snapshots from huggingface, as mentioned in Table~\ref{tab:links}. For Glide experiments, we use the official implementation and the released model snapshots. Following the recommendation, we use the refiner in Stable Diffusion XL and the upsampler in GLIDE. The diffusion steps and guidance scale hyperparameters we used are shown in Table~\ref{tab:DMParams}. For other hyperparameters, we follow the recommendations from Huggingface and GLIDE's official implementation. We use the prompt "An image of a {class\_name}." as the text guidance to generate the image and interpolate the generated image to the size of the target dataset (32 for CIFAR, 64 for TinyImageNet, and 224 for ImageNet-100).

\textbf{Samples from the generated datasets.}
Fig.~\ref{fig:samples} shows some samples from class ``n01558993'' (Robin) in the SDXL-In100 dataset along with the samples from the original dataset. We can notice a significant loss of diversity in the samples from SDXL-In100.

\subsection{Details about synthetic contamination simulation.}
\label{apdx:simulation}
As mentioned in Sec.~\ref{subsec:sim}, we generate synthetic twins of benchmark datasets and substitute a fixed portion $P$ of the original datasets with their synthetic counterparts. Similar to the class-wise distribution of Mix-C10/C100/Tiny, we also conduct the mixture class-wise. For datasets in setting \textbf{a)}, we assure that the contamination ratio in each class is also $P$. For datasets in setting \textbf{b)}, while maintaining a consistent class-wise contamination ratio, we also ensure that each individual synthetic model contributes 20\% of the contamination in each class.

\subsection{Task sequence.}
\label{apdx:taskseq}
In some work, the authors use a fixed task sequence for fair comparison. However, the final performance is largely affected by the task order. For fair comparison, we randomly assign the class to tasks and shuffle the sequence of tasks with 10 fixed random seeds. This ensures the evaluation is not biased to task difficulty.

\subsection{Data augmentation.}
\label{apdx:aug}
Data augmentation is effective in boosting the training of online continual learners. Methods may benefit differently from different augmentation intensities, and some methods may favor simpler augmentations instead of complicated ones. For a fair comparison, it is vital to ensure all methods are in their optimal performance. Thus, we introduce two different augmentation strategies:

\textbf{1. Partial strategy.} The partial augmentation is a weaker version of augmentation, consisting of random cropping with $p=0.5$, followed by random horizontal flip with $p=0.5$.

\textbf{2. Full strategy.} The full augmentation strategy is a stronger version of augmentation. The full augmentation strategy is a superset of its partial counterpart, which consists of random cropping, random horizontal flip, color jitter, and random grayscale. The parameters for color jitter are set to $(0.4, 0.4, 0.4, 0.1)$ with $p=0.8$. The probability of random grayscale is set to $p=0.2$.

We define the data augmentation strategy of each method with a hyperparameter search, as detailed in Appendix~\ref{apdx:hpsearch}.

\subsection{Hyperparameter search protocol.}
\label{apdx:hpsearch}
For hyperparameters in all of the methods (except DER++ on TinyImageNet), we conduct a hyperparameter search on the clean CIFAR-100 dataset with a memory size of 5,000, and apply the same hyperparameter to all of the other settings. The exhaustive list of the hyperparameter search is shown in Table~\ref{tab:hpsearch}.

\textbf{Special treatment for DER++.} DER++ encounters a catastrophic performance defect (close to 0) when trained on the TinyImageNet dataset using an optimizer with momentum. Thus, we applied another hyperparameter search for DER++ on TinyImageNet and found the SGD optimizer (without Momentum) gives reasonable performance. All the experiments of DER++ on TinyImageNet are conducted using these new hyperparameters.

\subsection{Hardware and computation.}
\begin{figure}[h]
    \centering
    \includegraphics[width=0.5\linewidth]{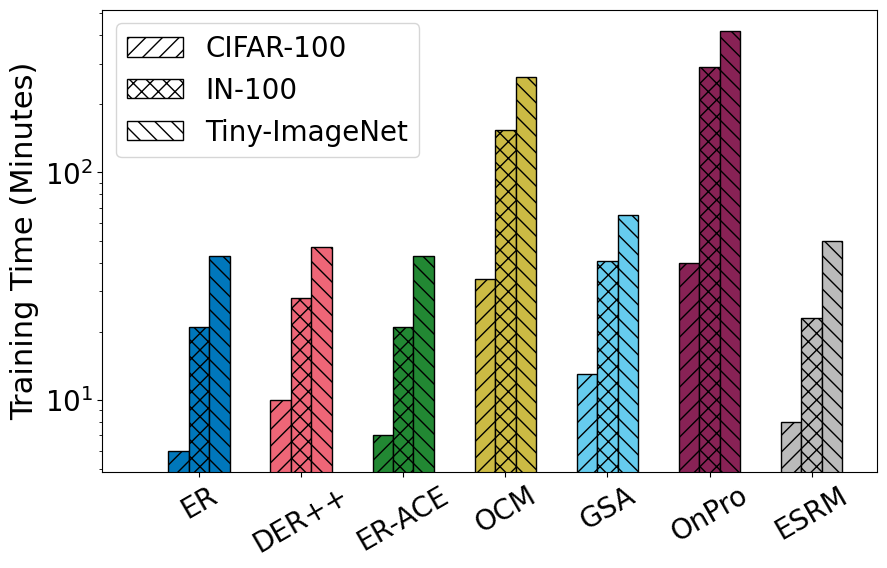}
    \caption{The average training time of each method trained on CIFAR-100 (M=5k), IN-100 (M=5k), and TinyImageNet (M=10k) dataset. For better readability, the values are plotted on the logarithm scale. The numbers are averaged from 10 runs.}
    \label{fig:compute}
\end{figure}

All of the experiments are conducted on NVIDIA A100 GPUs. The average training time of each method on CIFAR-100 (Memory size = 5k), ImageNet-100 (Memory size = 5k), and TinyImageNet (Memory size = 10k) is shown in Fig.~\ref{fig:compute}. The training efficiency is much faster than OCM and OnPro, while on par with the most efficient method.

\subsection{Useful source code links.}
For continual learning baselines, we use the codebase listed in Table~\ref{tab:links} to reimplement baseline methods. For image generation methods, we use the Diffuser library from Hugging Face for Stable Diffusion and VQDM experiments, and we use the codebase in Table~\ref{tab:links} for GLIDE.

\begin{table}[h]
    \centering
    \resizebox{0.75\linewidth}{!}{
    \begin{tabular}{ccc}
\toprule
Method & HP & Values \\
\midrule
\multirow{5}{*}{ER}& optimizer & [SGD, AdamW]\\
    & lr & [0.1, 0.05, 0.01, 0.005, 0.001, 0.0005, 0.0001] \\
    & weight decay & [0, 1e-4] \\
    & momentum (for SGD) & [0, 0.9] \\
    & aug. strat. & [partial, full] \\
\midrule
\multirow{7}{*}{DER++}& optimizer & [SGD, AdamW]\\
    & lr & [0.1, 0.05, 0.01, 0.005, 0.001, 0.0005, 0.0001] \\
    & weight decay & [0, 1e-4] \\
    & momentum (for SGD) & [0, 0.9] \\
    & aug. strat. & [partial, full] \\
    & alpha & [0.1, 0.2, 0.5, 1.0] \\
    & beta & [0.5, 1.0] \\
\midrule
\multirow{5}{*}{ER-ACE}& optimizer & [SGD, AdamW]\\
    & lr & [0.1, 0.05, 0.01, 0.005, 0.001, 0.0005, 0.0001] \\
    & weight decay & [0, 1e-4] \\
    & momentum (for SGD) & [0, 0.9] \\
    & aug. strat. & [partial, full] \\
\midrule
\multirow{4}{*}{OCM}& optimizer & [AdamW]\\
    & lr & [0.001] \\
    & weight decay & [1e-4] \\
    & aug. strat. & [partial, full] \\
\midrule
\multirow{5}{*}{GSA}& optimizer & [SGD, AdamW]\\
    & lr & [0.1, 0.05, 0.01, 0.005, 0.001, 0.0005, 0.0001] \\
    & weight decay & [0, 1e-4] \\
    & momentum (for SGD) & [0, 0.9] \\
    & aug. strat. & [partial, full] \\
\midrule
\multirow{5}{*}{OnPro}& optimizer & [SGD, AdamW]\\
    & lr & [0.1, 0.05, 0.01, 0.005, 0.001, 0.0005, 0.0001] \\
    & weight decay & [0, 1e-4] \\
    & momentum (for SGD) & [0, 0.9] \\
    & aug. strat. & [partial, full] \\
\midrule
\multirow{7}{*}{ESRM}& optimizer & [SGD, AdamW]\\
    & lr & [0.1, 0.05, 0.01, 0.005, 0.001, 0.0005, 0.0001] \\
    & weight decay & [0, 1e-4] \\
    & momentum (for SGD) & [0, 0.9] \\
    & aug. strat. & [partial, full] \\
    & $\lambda_1$ & [0.1, 0.2, 0.5, 1, 2, 5] \\
    & $\lambda_2$ & [0.1, 0.2, 0.5, 1, 2, 5] \\
\bottomrule
\end{tabular}
    }
    \vspace{5pt}
    \caption{Exhaustive list of hyperparameters searched on CIFAR-100.}
    \label{tab:hpsearch}
\end{table}

\begin{table}[h]
    \centering
    \resizebox{.6\linewidth}{!}{
    
\begin{tabular}{cc}
\toprule
Baseline & Links \\
\midrule
ER \& ER-ACE & \url{https://github.com/pclucas14/AML} \\
DER++ & \url{https://github.com/aimagelab/mammoth} \\
OCM & \url{https://github.com/gydpku/OCM} \\
GSA & \url{https://github.com/gydpku/GSA} \\
OnPro & \url{https://github.com/weilllllls/OnPro} \\
\midrule
Stable Diffusion \& VQDM & \url{https://github.com/huggingface/diffusers} \\
GLIDE & \url{https://github.com/openai/glide-text2im} \\
\bottomrule
\end{tabular}
    }
    \vspace{5pt}
    \caption{Baselines and their source code URLs.}
    \label{tab:links}
\end{table}
\vfill

\section*{NeurIPS Paper Checklist}

\begin{enumerate}

\item {\bf Claims}
    \item[] Question: Do the main claims made in the abstract and introduction accurately reflect the paper's contributions and scope?
    \item[] Answer: \answerYes{} 
    \item[] Justification: All the main claims made in both the abstract and introduction can accurately reflect the paper's contribution.
    \item[] Guidelines:
    \begin{itemize}
        \item The answer NA means that the abstract and introduction do not include the claims made in the paper.
        \item The abstract and/or introduction should clearly state the claims made, including the contributions made in the paper and important assumptions and limitations. A No or NA answer to this question will not be perceived well by the reviewers. 
        \item The claims made should match theoretical and experimental results, and reflect how much the results can be expected to generalize to other settings. 
        \item It is fine to include aspirational goals as motivation as long as it is clear that these goals are not attained by the paper. 
    \end{itemize}

\item {\bf Limitations}
    \item[] Question: Does the paper discuss the limitations of the work performed by the authors?
    \item[] Answer: \answerYes{} 
    \item[] Justification: The limitation section is included in the appendix.
    \item[] Guidelines:
    \begin{itemize}
        \item The answer NA means that the paper has no limitation while the answer No means that the paper has limitations, but those are not discussed in the paper. 
        \item The authors are encouraged to create a separate "Limitations" section in their paper.
        \item The paper should point out any strong assumptions and how robust the results are to violations of these assumptions (e.g., independence assumptions, noiseless settings, model well-specification, asymptotic approximations only holding locally). The authors should reflect on how these assumptions might be violated in practice and what the implications would be.
        \item The authors should reflect on the scope of the claims made, e.g., if the approach was only tested on a few datasets or with a few runs. In general, empirical results often depend on implicit assumptions, which should be articulated.
        \item The authors should reflect on the factors that influence the performance of the approach. For example, a facial recognition algorithm may perform poorly when image resolution is low or images are taken in low lighting. Or a speech-to-text system might not be used reliably to provide closed captions for online lectures because it fails to handle technical jargon.
        \item The authors should discuss the computational efficiency of the proposed algorithms and how they scale with dataset size.
        \item If applicable, the authors should discuss possible limitations of their approach to address problems of privacy and fairness.
        \item While the authors might fear that complete honesty about limitations might be used by reviewers as grounds for rejection, a worse outcome might be that reviewers discover limitations that aren't acknowledged in the paper. The authors should use their best judgment and recognize that individual actions in favor of transparency play an important role in developing norms that preserve the integrity of the community. Reviewers will be specifically instructed to not penalize honesty concerning limitations.
    \end{itemize}

\item {\bf Theory Assumptions and Proofs}
    \item[] Question: For each theoretical result, does the paper provide the full set of assumptions and a complete (and correct) proof?
    \item[] Answer: \answerNA{} 
    \item[] Justification: This paper does not include theoretical results.
    \item[] Guidelines:
    \begin{itemize}
        \item The answer NA means that the paper does not include theoretical results. 
        \item All the theorems, formulas, and proofs in the paper should be numbered and cross-referenced.
        \item All assumptions should be clearly stated or referenced in the statement of any theorems.
        \item The proofs can either appear in the main paper or the supplemental material, but if they appear in the supplemental material, the authors are encouraged to provide a short proof sketch to provide intuition. 
        \item Inversely, any informal proof provided in the core of the paper should be complemented by formal proofs provided in appendix or supplemental material.
        \item Theorems and Lemmas that the proof relies upon should be properly referenced. 
    \end{itemize}

    \item {\bf Experimental Result Reproducibility}
    \item[] Question: Does the paper fully disclose all the information needed to reproduce the main experimental results of the paper to the extent that it affects the main claims and/or conclusions of the paper (regardless of whether the code and data are provided or not)?
    \item[] Answer: \answerYes{} 
    \item[] Justification: The source code of our work, along with the result of the hyperparameter search are included in the supplementary material.
    \item[] Guidelines:
    \begin{itemize}
        \item The answer NA means that the paper does not include experiments.
        \item If the paper includes experiments, a No answer to this question will not be perceived well by the reviewers: Making the paper reproducible is important, regardless of whether the code and data are provided or not.
        \item If the contribution is a dataset and/or model, the authors should describe the steps taken to make their results reproducible or verifiable. 
        \item Depending on the contribution, reproducibility can be accomplished in various ways. For example, if the contribution is a novel architecture, describing the architecture fully might suffice, or if the contribution is a specific model and empirical evaluation, it may be necessary to either make it possible for others to replicate the model with the same dataset, or provide access to the model. In general. releasing code and data is often one good way to accomplish this, but reproducibility can also be provided via detailed instructions for how to replicate the results, access to a hosted model (e.g., in the case of a large language model), releasing of a model checkpoint, or other means that are appropriate to the research performed.
        \item While NeurIPS does not require releasing code, the conference does require all submissions to provide some reasonable avenue for reproducibility, which may depend on the nature of the contribution. For example
        \begin{enumerate}
            \item If the contribution is primarily a new algorithm, the paper should make it clear how to reproduce that algorithm.
            \item If the contribution is primarily a new model architecture, the paper should describe the architecture clearly and fully.
            \item If the contribution is a new model (e.g., a large language model), then there should either be a way to access this model for reproducing the results or a way to reproduce the model (e.g., with an open-source dataset or instructions for how to construct the dataset).
            \item We recognize that reproducibility may be tricky in some cases, in which case authors are welcome to describe the particular way they provide for reproducibility. In the case of closed-source models, it may be that access to the model is limited in some way (e.g., to registered users), but it should be possible for other researchers to have some path to reproducing or verifying the results.
        \end{enumerate}
    \end{itemize}

\item {\bf Open access to data and code}
    \item[] Question: Does the paper provide open access to the data and code, with sufficient instructions to faithfully reproduce the main experimental results, as described in supplemental material?
    \item[] Answer: \answerYes{} 
    \item[] Justification: The code is included in the supplementary material. 
    \item[] Guidelines:
    \begin{itemize}
        \item The answer NA means that paper does not include experiments requiring code.
        \item Please see the NeurIPS code and data submission guidelines (\url{https://nips.cc/public/guides/CodeSubmissionPolicy}) for more details.
        \item While we encourage the release of code and data, we understand that this might not be possible, so “No” is an acceptable answer. Papers cannot be rejected simply for not including code, unless this is central to the contribution (e.g., for a new open-source benchmark).
        \item The instructions should contain the exact command and environment needed to run to reproduce the results. See the NeurIPS code and data submission guidelines (\url{https://nips.cc/public/guides/CodeSubmissionPolicy}) for more details.
        \item The authors should provide instructions on data access and preparation, including how to access the raw data, preprocessed data, intermediate data, and generated data, etc.
        \item The authors should provide scripts to reproduce all experimental results for the new proposed method and baselines. If only a subset of experiments are reproducible, they should state which ones are omitted from the script and why.
        \item At submission time, to preserve anonymity, the authors should release anonymized versions (if applicable).
        \item Providing as much information as possible in supplemental material (appended to the paper) is recommended, but including URLs to data and code is permitted.
    \end{itemize}

\item {\bf Experimental Setting/Details}
    \item[] Question: Does the paper specify all the training and test details (e.g., data splits, hyperparameters, how they were chosen, type of optimizer, etc.) necessary to understand the results?
    \item[] Answer: \answerYes{} 
    \item[] Justification: The main experimental setting is included in the paper, and all of the details are included in the source code in the supplementary material.
    \item[] Guidelines:
    \begin{itemize}
        \item The answer NA means that the paper does not include experiments.
        \item The experimental setting should be presented in the core of the paper to a level of detail that is necessary to appreciate the results and make sense of them.
        \item The full details can be provided either with the code, in appendix, or as supplemental material.
    \end{itemize}

\item {\bf Experiment Statistical Significance}
    \item[] Question: Does the paper report error bars suitably and correctly defined or other appropriate information about the statistical significance of the experiments?
    \item[] Answer: \answerYes{} 
    \item[] Justification: Error bars are reported in the experiments. 
    \item[] Guidelines:
    \begin{itemize}
        \item The answer NA means that the paper does not include experiments.
        \item The authors should answer "Yes" if the results are accompanied by error bars, confidence intervals, or statistical significance tests, at least for the experiments that support the main claims of the paper.
        \item The factors of variability that the error bars are capturing should be clearly stated (for example, train/test split, initialization, random drawing of some parameter, or overall run with given experimental conditions).
        \item The method for calculating the error bars should be explained (closed form formula, call to a library function, bootstrap, etc.)
        \item The assumptions made should be given (e.g., Normally distributed errors).
        \item It should be clear whether the error bar is the standard deviation or the standard error of the mean.
        \item It is OK to report 1-sigma error bars, but one should state it. The authors should preferably report a 2-sigma error bar than state that they have a 96\% CI, if the hypothesis of Normality of errors is not verified.
        \item For asymmetric distributions, the authors should be careful not to show in tables or figures symmetric error bars that would yield results that are out of range (e.g. negative error rates).
        \item If error bars are reported in tables or plots, The authors should explain in the text how they were calculated and reference the corresponding figures or tables in the text.
    \end{itemize}

\item {\bf Experiments Compute Resources}
    \item[] Question: For each experiment, does the paper provide sufficient information on the computer resources (type of compute workers, memory, time of execution) needed to reproduce the experiments?
    \item[] Answer: \answerYes{} 
    \item[] Justification: Detailed information is included in the appendix.
    \item[] Guidelines:
    \begin{itemize}
        \item The answer NA means that the paper does not include experiments.
        \item The paper should indicate the type of compute workers CPU or GPU, internal cluster, or cloud provider, including relevant memory and storage.
        \item The paper should provide the amount of compute required for each of the individual experimental runs as well as estimate the total compute. 
        \item The paper should disclose whether the full research project required more compute than the experiments reported in the paper (e.g., preliminary or failed experiments that didn't make it into the paper). 
    \end{itemize}
    
\item {\bf Code Of Ethics}
    \item[] Question: Does the research conducted in the paper conform, in every respect, with the NeurIPS Code of Ethics \url{https://neurips.cc/public/EthicsGuidelines}?
    \item[] Answer: \answerYes{} 
    \item[] Justification: The Code of Ethics is fully respected and obeyed in our research. 
    \item[] Guidelines:
    \begin{itemize}
        \item The answer NA means that the authors have not reviewed the NeurIPS Code of Ethics.
        \item If the authors answer No, they should explain the special circumstances that require a deviation from the Code of Ethics.
        \item The authors should make sure to preserve anonymity (e.g., if there is a special consideration due to laws or regulations in their jurisdiction).
    \end{itemize}

\item {\bf Broader Impacts}
    \item[] Question: Does the paper discuss both potential positive societal impacts and negative societal impacts of the work performed?
    \item[] Answer: \answerNA{} 
    \item[] Justification: There is no societal impact of the work performed.
    \item[] Guidelines:
    \begin{itemize}
        \item The answer NA means that there is no societal impact of the work performed.
        \item If the authors answer NA or No, they should explain why their work has no societal impact or why the paper does not address societal impact.
        \item Examples of negative societal impacts include potential malicious or unintended uses (e.g., disinformation, generating fake profiles, surveillance), fairness considerations (e.g., deployment of technologies that could make decisions that unfairly impact specific groups), privacy considerations, and security considerations.
        \item The conference expects that many papers will be foundational research and not tied to particular applications, let alone deployments. However, if there is a direct path to any negative applications, the authors should point it out. For example, it is legitimate to point out that an improvement in the quality of generative models could be used to generate deepfakes for disinformation. On the other hand, it is not needed to point out that a generic algorithm for optimizing neural networks could enable people to train models that generate Deepfakes faster.
        \item The authors should consider possible harms that could arise when the technology is being used as intended and functioning correctly, harms that could arise when the technology is being used as intended but gives incorrect results, and harms following from (intentional or unintentional) misuse of the technology.
        \item If there are negative societal impacts, the authors could also discuss possible mitigation strategies (e.g., gated release of models, providing defenses in addition to attacks, mechanisms for monitoring misuse, mechanisms to monitor how a system learns from feedback over time, improving the efficiency and accessibility of ML).
    \end{itemize}
    
\item {\bf Safeguards}
    \item[] Question: Does the paper describe safeguards that have been put in place for responsible release of data or models that have a high risk for misuse (e.g., pretrained language models, image generators, or scraped datasets)?
    \item[] Answer: \answerNA{} 
    \item[] Justification: The paper does not pose such risks.
    \item[] Guidelines:
    \begin{itemize}
        \item The answer NA means that the paper poses no such risks.
        \item Released models that have a high risk for misuse or dual-use should be released with necessary safeguards to allow for controlled use of the model, for example by requiring that users adhere to usage guidelines or restrictions to access the model or implementing safety filters. 
        \item Datasets that have been scraped from the Internet could pose safety risks. The authors should describe how they avoided releasing unsafe images.
        \item We recognize that providing effective safeguards is challenging, and many papers do not require this, but we encourage authors to take this into account and make a best faith effort.
    \end{itemize}

\item {\bf Licenses for existing assets}
    \item[] Question: Are the creators or original owners of assets (e.g., code, data, models), used in the paper, properly credited and are the license and terms of use explicitly mentioned and properly respected?
    \item[] Answer: \answerYes{} 
    \item[] Justification: We have cited the used assets in the appendix. The licenses of existing assets are properly respected.
    \item[] Guidelines:
    \begin{itemize}
        \item The answer NA means that the paper does not use existing assets.
        \item The authors should cite the original paper that produced the code package or dataset.
        \item The authors should state which version of the asset is used and, if possible, include a URL.
        \item The name of the license (e.g., CC-BY 4.0) should be included for each asset.
        \item For scraped data from a particular source (e.g., website), the copyright and terms of service of that source should be provided.
        \item If assets are released, the license, copyright information, and terms of use in the package should be provided. For popular datasets, \url{paperswithcode.com/datasets} has curated licenses for some datasets. Their licensing guide can help determine the license of a dataset.
        \item For existing datasets that are re-packaged, both the original license and the license of the derived asset (if it has changed) should be provided.
        \item If this information is not available online, the authors are encouraged to reach out to the asset's creators.
    \end{itemize}

\item {\bf New Assets}
    \item[] Question: Are new assets introduced in the paper well documented and is the documentation provided alongside the assets?
    \item[] Answer: \answerYes{} 
    \item[] Justification: We have a Readme file along with our source code.
    \item[] Guidelines:
    \begin{itemize}
        \item The answer NA means that the paper does not release new assets.
        \item Researchers should communicate the details of the dataset/code/model as part of their submissions via structured templates. This includes details about training, license, limitations, etc. 
        \item The paper should discuss whether and how consent was obtained from people whose asset is used.
        \item At submission time, remember to anonymize your assets (if applicable). You can either create an anonymized URL or include an anonymized zip file.
    \end{itemize}

\item {\bf Crowdsourcing and Research with Human Subjects}
    \item[] Question: For crowdsourcing experiments and research with human subjects, does the paper include the full text of instructions given to participants and screenshots, if applicable, as well as details about compensation (if any)? 
    \item[] Answer: \answerNA{} 
    \item[] Justification: The paper does not involve crowdsourcing or research with human subjects.
    \item[] Guidelines:
    \begin{itemize}
        \item The answer NA means that the paper does not involve crowdsourcing nor research with human subjects.
        \item Including this information in the supplemental material is fine, but if the main contribution of the paper involves human subjects, then as much detail as possible should be included in the main paper. 
        \item According to the NeurIPS Code of Ethics, workers involved in data collection, curation, or other labor should be paid at least the minimum wage in the country of the data collector. 
    \end{itemize}

\item {\bf Institutional Review Board (IRB) Approvals or Equivalent for Research with Human Subjects}
    \item[] Question: Does the paper describe potential risks incurred by study participants, whether such risks were disclosed to the subjects, and whether Institutional Review Board (IRB) approvals (or an equivalent approval/review based on the requirements of your country or institution) were obtained?
    \item[] Answer: \answerNA{} 
    \item[] Justification: The paper does not involve crowdsourcing or research with human subjects.
    \item[] Guidelines:
    \begin{itemize}
        \item The answer NA means that the paper does not involve crowdsourcing nor research with human subjects.
        \item Depending on the country in which research is conducted, IRB approval (or equivalent) may be required for any human subjects research. If you obtained IRB approval, you should clearly state this in the paper. 
        \item We recognize that the procedures for this may vary significantly between institutions and locations, and we expect authors to adhere to the NeurIPS Code of Ethics and the guidelines for their institution. 
        \item For initial submissions, do not include any information that would break anonymity (if applicable), such as the institution conducting the review.
    \end{itemize}

\end{enumerate}


\end{document}